\documentclass[journal]{IEEEtran}
\usepackage{amsmath,amsfonts}
\usepackage{algorithmic}
\usepackage{algorithm}
\usepackage{array}
\usepackage{textcomp}
\usepackage{stfloats}
\usepackage{url}
\usepackage{booktabs}
\usepackage{multirow}
\usepackage{verbatim}
\usepackage{graphicx}
\usepackage{soul, color, xcolor}
\definecolor{citecolor}{RGB}{66,168,235}
\definecolor{linkcolor}{RGB}{255,0,0}
\definecolor{urlcolor}{RGB}{0,0,0}
\definecolor{greencolor}{RGB}{0.204,0.659,0.325}
\usepackage{hyperref}
\hypersetup{colorlinks=true,citecolor=citecolor,linkcolor=linkcolor,urlcolor=urlcolor}
\usepackage{stfloats}
\soulregister{\ref}7
\soulregister{\cite}7
\soulregister\citep7 
\soulregister\citet7 
\soulregister\ref7 
\soulregister\pageref7 

\begin{document}
\title{BD-MSA: Body decouple VHR Remote Sensing Image Change Detection method guided by multi-scale feature information aggregation}
\author{Yonghui~Tan,
        Xiaolong~Li,
        Yishu~Chen,
        and~Jinquan~Ai
\thanks{Yonghui Tan, Xiaolong Li and Jinquan Ai was with the Key Laboratory of Mine Environmental Monitoring and Improving around Poyang Lake, Ministry of Natural Resources, East China University of Technology, Nanchang 330013, China (email: lixiaolong@ecut.edu.cn, cv\underline{~}tyh@ecut.edu.cn), jinquan@ecut.edu.cn (\emph{Corresponding author: Xiaolong Li}).}
\thanks{Yishu Chen are with Ningbo Alatu Digital Technology Co., Ltd, Ningbo 315000, China (email: 2817161223@qq.com).}
}


\maketitle

\begin{abstract}
The purpose of remote sensing image change detection (RSCD) is to detect differences between bi-temporal images taken at the same place. Deep learning has been extensively used to RSCD tasks, yielding significant results in terms of result recognition. However, due to the shooting angle of the satellite, the impacts of thin clouds, and certain lighting conditions, the problem of fuzzy edges in the change region in some remote sensing photographs cannot be properly handled using current RSCD algorithms. To solve this issue, we proposed \textit{a Body Decouple Multi-Scale by fearure Aggregation change detection} (BD-MSA), a novel model that collects both global and local feature map information in the channel and space dimensions of the feature map during the training and prediction phases. This approach allows us to successfully extract the change region's boundary information while also divorcing the change region's main body from its boundary. Numerous studies have shown that the assessment metrics and evaluation effects of the model described in this paper on the publicly available datasets DSIFN-CD, S2Looking and WHU-CD are the best when compared to other models.
\end{abstract}


\section{Introduction}
\label{sec:introduction}

\IEEEPARstart{C}{hange} detection (CD) is a technique for determining whether a change has occurred at the same area by examining images of that location at different times~\cite{singh1989review,bai2023deep,shafique2022deep}. Binary change detection is a popular technique that analyzes information between two images to determine whether a pixel in one image has changed. It then categorizes the pixels in the image as either changed or unchanged. One of the fundamental and essential issues in remote sensing is the interpretation of very-high-resolution (VHR) remote sensing (RS) images. VHR remote sensing image change detection (RSCD) is useful for a variety of remote sensing applications, including urban land use analysis~\cite{onur2009change,tariq2023modeling,ray2023quantitative}, building detection~\cite{li2022review,huang2016building,maltezos2017deep}, deforestation monitoring~\cite{kaselimi2022vision,solorzano2023deforestation}, urban planning~\cite{kennedy2009remote,gomroki2023stcd}, urban sprawl analysis~\cite{du2012fusion,arif2023dynamics}, disaster assessment~\cite{zheng2021building,pedzisai2023novel}, and so on. All of the aforementioned are necessary for local governments to effectively manage their local urban development, and precise and effective RSCD procedures enable cities to be assessed and planned for in order to minimize or prevent adverse effects.

The challenge in RSCD is capturing the connections between regions of interest between bi-temporal images while disregarding interference from other regions. At the same time, several irritating elements such as seasonality in the bi-temporal and image quality issues such as noise and contrast are not of importance and should be ignored when performing CD.

\begin{figure}[t]
	\centering
	\includegraphics[width=1\linewidth]{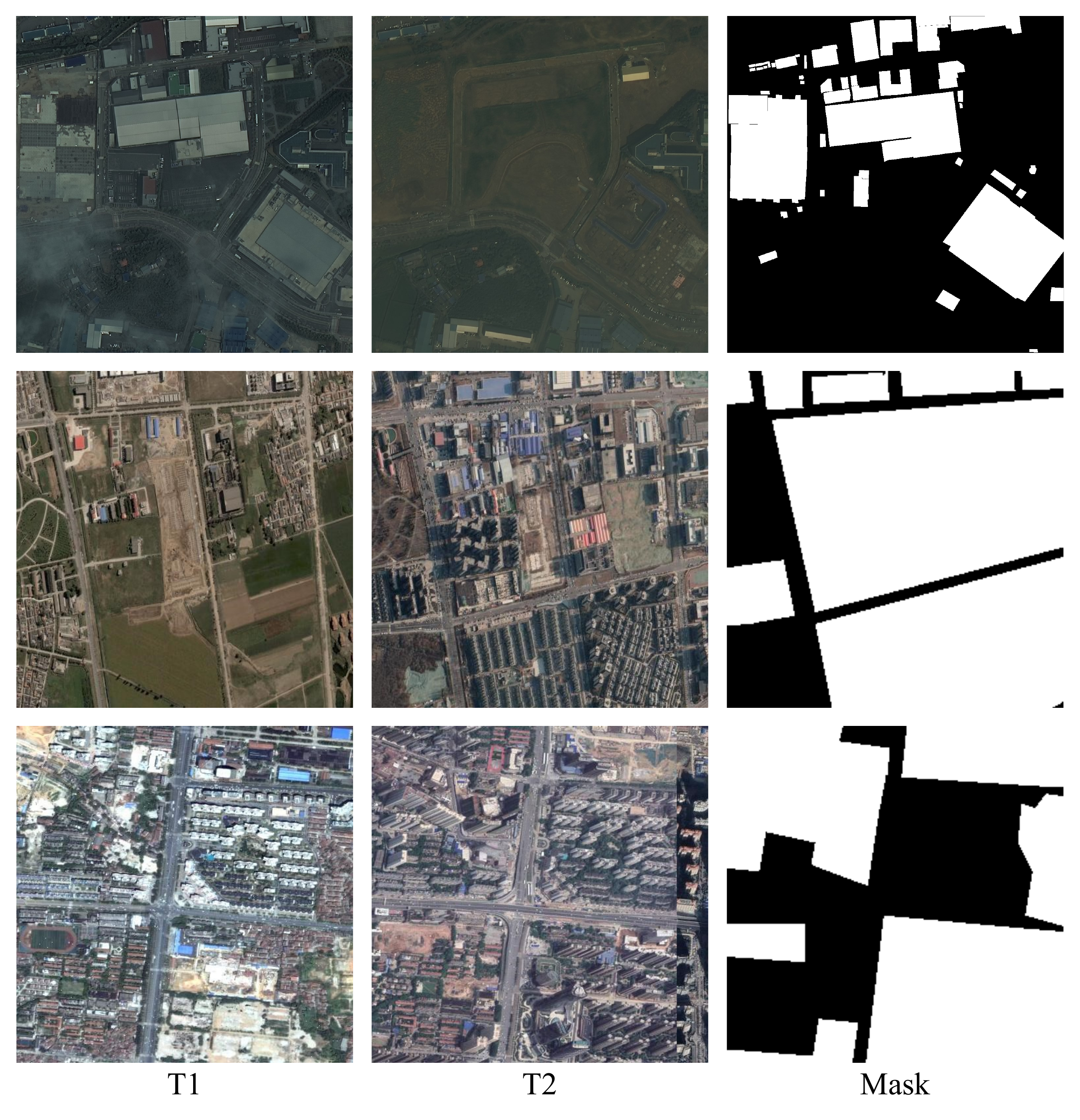}
	\caption{A section of the images in the DSIFN-CD and S2Looking, with the first column representing the pre-change images, the second representing the post-change images, and the third representing the change mask. The photos in the figure's top row are from S2Looking, while those in the second and third rows are from DSIFN-CD.}
	\label{fig:Blurred_edges}
\end{figure}

The two primary streams of CD in RS images are the traditional method and the deep learning method, which has gained popularity in the last decade. Several strategies for detecting changes in RS images have been proposed before using deep learning to RS images on a broad scale. Coppin and Bauer~\cite{coppin1996digital} employed a pixel-based change detection method for RSCD, which detects changes in gray values or colors by comparing images from two points in time pixel by pixel. Deng et al.~\cite{deng2008pca} discovered and quantified land use change using PCA and a hybrid classifier that includes both unsupervised and supervised classification. He et al.~\cite{he2011detecting} combined texture change information with standard spectral-based change vector analysis (CVA), resulting in integrated spectral and texture change information. Wu et al.~\cite{wu2013slow} used slow feature analysis to isolate the most time-undeformed section of a multi-temporal image and migrate it to a new feature space, effectively concealing the image's unaltered pixels. 

Although these techniques have produced good results, they have certain drawbacks because they rely on standard image processing:
\begin{itemize}
	\item Traditional techniques often necessitate the manual design of features, which may necessitate domain expertise and experience;
	\item When dealing with complicated sceneries, varied lighting conditions, and multi-category changes, traditional approaches have rather weak generalization capacity;
	\item For supervised learning, traditional methods often necessitate enormous amounts of manually labeled data.
\end{itemize}

Deep learning is a technique that has evolved tremendously quickly in the previous decade, and deep learning-based computer vision has achieved exceptional performance in RSCD tasks because to CNNs' robust feature extraction capabilities. Deep learning computer vision based RSCD techniques can be divided into three categories based on the structure of these models: pure convolution based, attention mechanism based, and Transformer based. The above can be categorized as 1)  Fully Convolutional (FC-EF, FC-Siam-Di, FC-Siam-Conc)~\cite{daudt2018fully}, Improved UNet++~\cite{peng2019end}, IFNet~\cite{zhang2020deeply}, CDNet~\cite{chen2021adversarial}, DTCDSCN~\cite{liu2020building}, TINY-CD~\cite{codegoni2023tinycd}, these models simply extract features from RS images using CNNs, which make it difficult to capture long-term dependencies across images and may be insensitive to complicated scene changes. 2) MSPSNet~\cite{guo2021deep}, DSAMNet~\cite{shi2021deeply}, HANet~\cite{han2023hanet}, STANet~\cite{chen2020spatial}, SNUNet~\cite{fang2021snunet}, ADS-Net~\cite{wang2021ads}, DARNet~\cite{li2022densely}, SRCDNet~\cite{liu2021super}, TFI-GR~\cite{li2022remote}, the strategies described above boost the model's sensitivity to crucial regions and help improve change detection accuracy, but it is challenging to collect global information in bi-temporal images. 3) BIT~\cite{chen2021remote}, ChangeFormer~\cite{bandara2022transformer}, RSP-BIT~\cite{wang2022empirical}, SwinSUNet~\cite{zhang2022swinsunet}, MTCNet~\cite{wang2022cbam}, TransUNetCD~\cite{li2022transunetcd}, DMATNet~\cite{song2022remote}, FTN~\cite{yan2022fully}, AMTNet~\cite{liu2023attention}, Hybrid-transcd~\cite{ke2022hybrid}, when compared to traditional convolutional approaches, Transformer can handle long-range relationships better, but its ability to extract local contextual information is poor and computationally expensive.

Although the methods described above produced outstanding results in the RSCD task, they have certain flaws. Because of their narrow local perceptual domain and susceptibility to spatial fluctuations, pure convolution-based approaches have limited degrees of feature extraction for RSCD. Second, when executing RSCD, the approach based on the attention mechanism can only take into consideration the local information in the feature map and cannot aggregate the global information. Third, the Transformer-based solution lacks the link between contexts in the details, and the arithmetic need is excessively large.

It is worth mentioning that the changing camera angles for different time phases, as well as the fact that most RS photographs are not shot at an angle perpendicular to the ground, result in shadows on various features in the enormous number of RS images. Furthermore, thin clouds appear in some RS photographs as a result of meteorological conditions. As illustrated in Fig.~\ref{fig:Blurred_edges}, the majority of the buildings in the image are inclined and cast long shadows on the ground, and thin clouds can be seen in some of the photographs. When executing RSCD, the margins of the change region are likely to get blurred due to the aforementioned issue. As a result, we prefer to address this issue during model training.

Targeting the two aforementioned primary issues—that is, the inability of current RSCD methods to effectively aggregate global and local feature information simultaneously and the blurring of change region edges as a result of feature shadowing in remote sensing images—we proposed \textbf{BD-MSA}, a model that can simultaneously aggregate global and local information in multi-scale feature maps and decouple the change region's center from its edges during training.

The contributions of this paper are as follows:
\begin{itemize}
	\item [(1)] The Overall Feature Aggregation Module (OFAM), which we proposed in this paper, is a technique that can simultaneously aggregated global and local information in both channel and spatial dimensions. It can adapt feature information at different scales in the backbone part while effectively increasing the model's accuracy;
	\item [(2)] Given the large difference in recognition accuracy between the main body and the edge of the changing region in the RSCD task, this paper designs a Decouple Module in the prediction head part that can effectively separate the main body and the edge of the changing region, and the experimental results show that using this module improves the model's recognition accuracy for the edge;
	\item [(3)] Since the MixFFN module in SegFormer can capture intricate feature representations in the network, this paper presents the module in the network decoder, enhancing the feature extraction and generalization capabilities of the model;
	\item [(4)] Extensive studies show that the technique presented in this work outperforms existing models on the public datasets DSIFN-CD and S2Looking, achieving the SOTA (state-of-the-art) performance.
\end{itemize}

The rest of the paper is structured as follows. The prior approaches are introduced in section~\ref{sec:related work}. The model's detail described in this paper is introduced in Section~\ref{sec:BD-MSA}. Section~\ref{sec:experiments} conducts experiments to compare this paper's method with related methods. The discussion is shown in Section~\ref{sec:discussion}. This paper is summarized in Section~\ref{sec:conclusion}.

\section{Related Work}
\label{sec:related work}
In this section, we present an overview of existing RSCD works, including: pure convolution-based, attention mechanism-based, and Transformer-based.

\subsection{Pure Convolutional-based Model}
Deep CNNs have achieved amazing performance in the field of computer vision~\cite{islam2016application} due to their powerful feature extraction capabilities. RS image interpretation is essentially an image processing in which deep learning plays an important role such as image classification~\cite{maggiori2016convolutional}, object detection~\cite{cheng2016survey,deng2018multi}, semantic segmentation~\cite{kemker2018algorithms,yuan2021review}, and change detection~\cite{2023Global}.

In the field of RSCD, the first attempt to use fully convolutional networks is the work of~\cite{daudt2018fully}, it devided into three methods namely FC-EF, FC-Siam-Di, and FC-Siam-Conc, it proposes a CD architecture for the Siamese Network and demonstrates that this architecture is effective; In~\cite{peng2019end}, an improved UNet++~\cite{zhou2018unet++} has been proposed, it adopts the MSOF strategy, which can effectively combine multi-scale information and help to detect objects with large size and scale variations on VHR RS images; Zhang et al.~\cite{zhang2020deeply} proposed a depth-supervised image fusion network for CD in high-resolution bi-temporal RS images, which combines the attention module and depth supervision to provide an effective new way for CD in RS images. For better industrial applications, Andrea et al.~\cite{codegoni2023tinycd} proposed TINY-CD, which employs the Siamese U-Net architecture and an innovative Mixed and Attention Masking Block (MAMB) to achieve better performance than existing models while being smaller in size.

\subsection{Attention Mechanism-based Model}
Attention mechanisms were first introduced in the context of natural language processing~\cite{vaswani2017attention}. Later, computer vision researchers presented attentional processes that could be applied to images~\cite{woo2018cbam,hu2018squeeze,li2022contextual}.

One can apply attention techniques in the field of RSCD, just like in most other computer vision jobs. In order to address issues like illumination noise and scale variations in aerial image change detection, Shi et al.~\cite{shi2021deeply} introduced a deeply supervised attentional metric network for remote sensing change detection, this network incorporates a metric learning module and a convolutional block attentional module (CBAM) to enhance feature differentiation; In order to increase detection accuracy, Guo et al.~\cite{guo2021deep} suggested a deep multiscale twin network for RSCD. This network is based on a deep multiscale twin neural network and incorporates a self-attention module and a parallel convolutional structure. Li et al.~\cite{li2022densely} proposed a dense attention refinement network that combines dense hopping connections, a hybrid attention module that combines a channel attention module and a spatial-temporal attention module, and a recursive refinement module to effectively improve the accuracy of CD in high-resolution bi-temporal RS images. In order to overcome the resolution disparity between bi-temporal images, Liu et al.~\cite{liu2021super} created SRCDNet, which learns super-resolution images using adversarial learning and enriches multiscale features with a stacked attention module made up of five CBAMs.

Even though attention-based RSCD is more adept at identifying local contextual information from bi-temporal RS images, it is less effective at capturing the global information.

\subsection{Transformer-based Model}
Transformer is crucial to RSCD because of its potent global feature extraction capacity. For the first time, BIT~\cite{chen2021remote} brings Transformer, which effectively describes context in the spatial-temporal domain to the RSCD domain. In order to model context and improve features, BIT converts the input image into a limited set of high-level semantic tokens using Transformer encoders and decoders; Based on BIT, RSP-BIT~\cite{wang2022empirical} primarily focuses on using Remote Sensing Pretraining (RSP) to analyze aerial images. It has been observed that RSP enhances performance on the scene identification test and helps comprehend the semantics related to RS; By fusing a multiscale Transformer with a CBAM, Wang et al.~\cite{wang2022cbam} developed MTCNet. It creates a multiscale module to create the multiscale Transformer after extracting the bi-temporal image features using the Transformer module.

While Transformer performs RSCD tasks effectively in terms of global information extraction, its huge number of parameters makes prediction more time-consuming, and it struggles to extract the semantics across local contexts.

\section{Methodology}
\label{sec:BD-MSA}
In this section, we proposed BD-MSA, a novel approach in which we first provide a brief overview of the general structure, followed by a full description of the modules in our approach in each subsection.

\subsection{Overall Structure}
The siamese network is presently a commonly utilized structure in RSCD, which uses two weight-sharing Backbone in the feature extraction phase to extract features from the input. In BD-MSA, we feed $I=\left\{I_1,I_2\right\}$ into the CNN Backbone to extract the respective deep features of the bi-temporal images, which are then sent successively through the Decouple Decoder and the Prediction Mask, and the output is compared with the Mask. 
\begin{figure*}[h]
	\centering
	\includegraphics[width=1\linewidth]{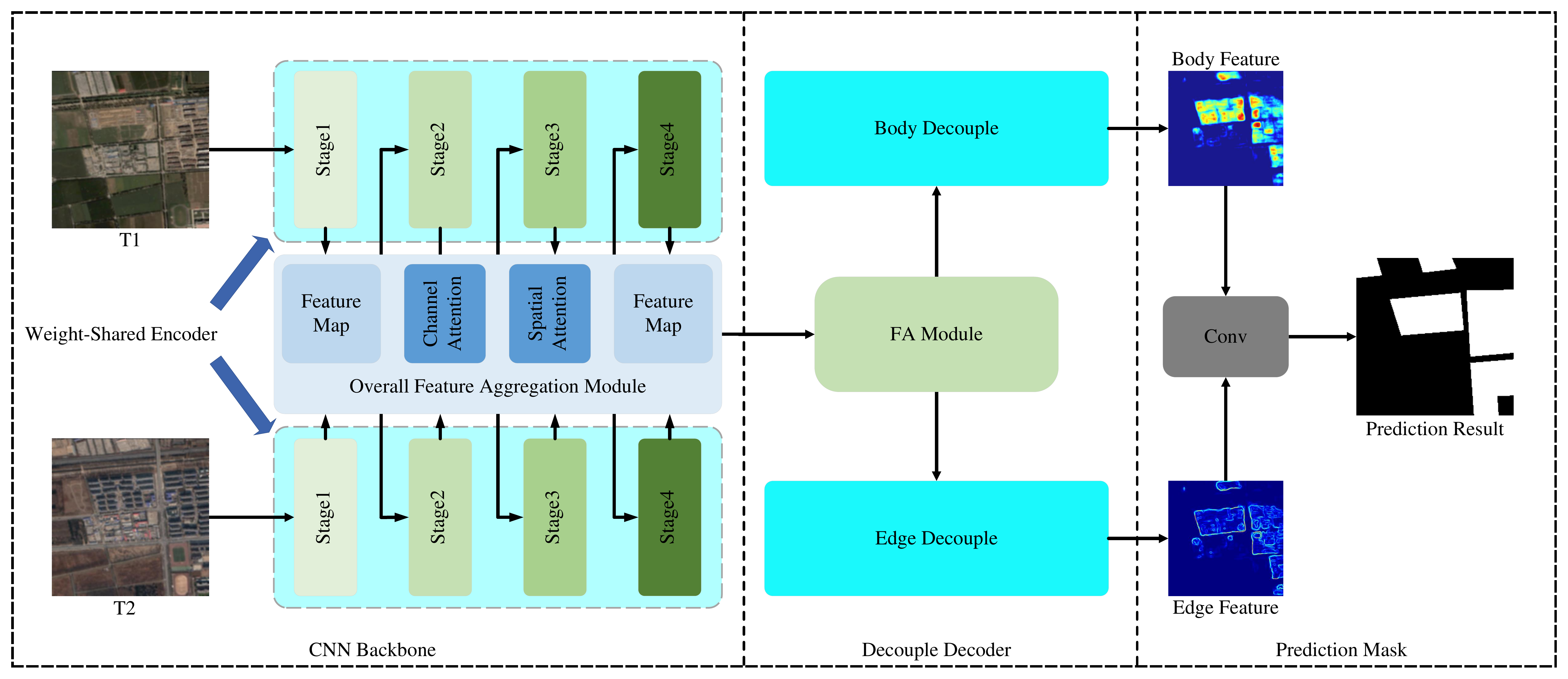}
	\caption{Schematic diagram of BD-MSA.}
	\label{fig:overall}
\end{figure*}

Fig.~\ref{fig:overall} depicts the general architecture diagram of BD-MSA. The diagram is divided into three primary sections: CNN Backbone, Decouple Decoder, and Prediction Mask. The following equation can illustrate the model training process:
\begin{equation}
	\rm \hat{Y}=Predict\left( Decoder\left( Backbone\left\{ I_1,I_2 \right\} \right) \right) 
	\label{eq:overall}
\end{equation}
where Backbone, Decoder, and Predict represent different parts of the model diagram, $\rm \hat{Y}$ represents the training result graph, and $\rm I_1$, $\rm I_2$ represent the input bi-temporal images. In Algorithm~\ref{algorithm}, we have expressed the model training procedure as pseudo-code to help the reader comprehend.
\begin{algorithm}[h]
	\caption{Inference of BD-MSA for Change Detection}
	\label{algorithm}
	\begin{algorithmic}[1]
		\REQUIRE $\textbf{I}=\left\lbrace \left( \textbf{I}^{1}, \textbf{I}^{2} \right) \right\rbrace $ (a pair of bi-temporal image)
		\ENSURE $\textbf{M}$ (a prediction change mask)
		\STATE // step1: extract high-level features by MiT backbone and OAFM
		\FOR{$i~in~\left\lbrace 1, 2\right\rbrace $}
		\FOR{$n~in~\left\lbrace 1, 2, 3, 4\right\rbrace $}
		\STATE $\textbf{MiT}^{i}_{n}={\rm MiT\underline{~}Backbone}(\textbf{T}^{i}) $
		\STATE $\textbf{F}^{i}_{n}={\rm OAFM}(\textbf{MiT}^{i}_{n}) $
		\ENDFOR
		\ENDFOR
		\STATE // step2: Concat high-level feature to FA Module
		\STATE $\textbf{F}_{FA}={\rm FA\underline{~}Module}(\textbf{F}^{1}_{4},\textbf{F}^{2}_{4})$
		\STATE // step3: Decoupling $\textbf{F}_{FA}$ into $\textbf{F}_{body}$ and $\textbf{F}_{edge}$ by Body Decouple and Edge Decouple
		\STATE $\textbf{F}_{body}={\rm Body\underline{~}Decouple}(\textbf{F}_{FA})$
		\STATE $\textbf{F}_{edge}={\rm Edge\underline{~}Decouple}(\textbf{F}_{FA})$
		\STATE $\textbf{M}={\rm Conv}({\rm Concat}(\textbf{F}_{body}, \textbf{F}_{edge}))$
	\end{algorithmic}
\end{algorithm}

\subsection{Overall Feature Aggregation Module (OFAM)}
\begin{figure*}[h]
	\centering
	\includegraphics[width=0.8\linewidth]{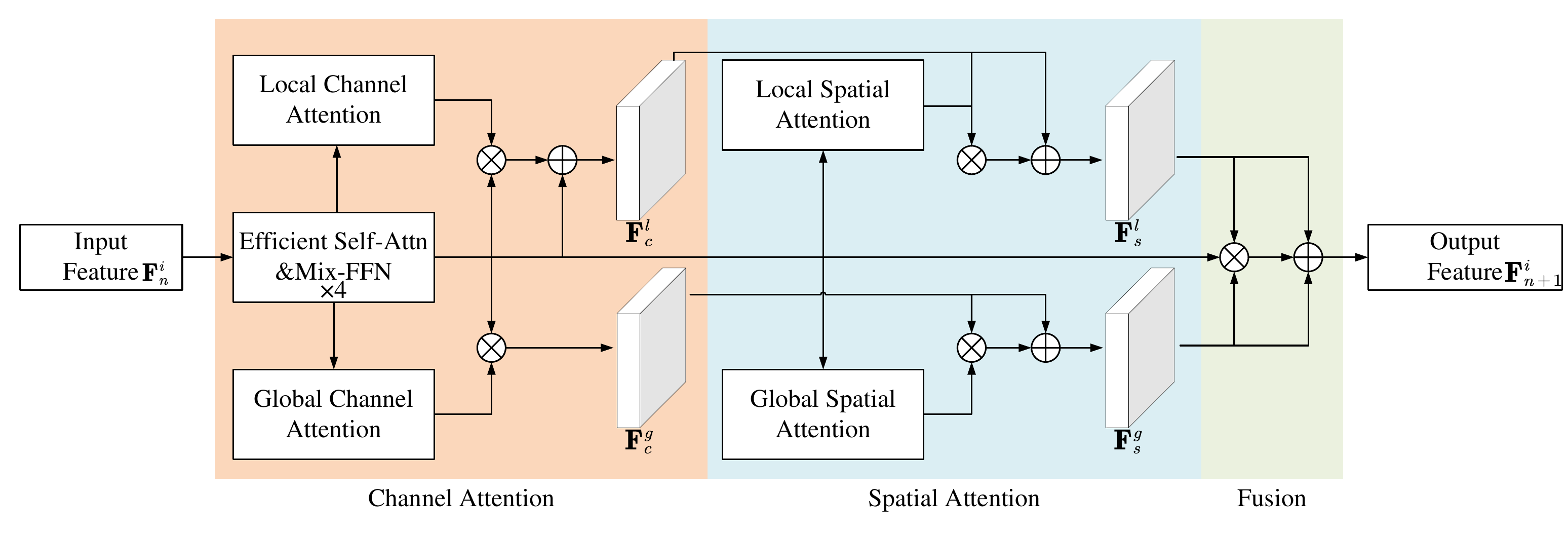}
	\caption{The graphic depicts our OFAM, which is separated into three major portions, Channel Attention, Spatial Attention, and Fusion, which are distinguished by various colored backgrounds.}
	\label{fig:OFAM}
\end{figure*}

In the feature extraction section, we utilize a feature decoder with shared weights to send the input diachronic phase through two identical Backbones with the same weights during the training process. The Backbone we designed, in particular, can be divided into four stages, in which the input features are first made to pass through the MiT~\cite{xie2021segformer}, because of its greater success in the field of semantic segmentation in recent years, and the output results are then pass through the OFAM while extracting both local and global features in the channel dimension and spatial dimension of the feature map. Fig.~\ref{fig:OFAM} depicts the OFAM module that we designed.

Our designed OFAM is divided into three parts. First, we divide the output of MiT in channel dimension into two branches, one of which, Local Channel Attention, is used to extract local features and the other branch, Global Channel Attention, is used to extract global features; the related computational formula is as follows:
\begin{equation}
	\mathbf{F}_{c}^{l}=\mathrm{MiT}\left( \mathbf{F}_{n}^{i} \right) +\mathrm{LCA}\left( \mathrm{MiT}\left( \mathbf{F}_{n}^{i} \right) \right) \times \mathrm{MiT}\left( \mathbf{F}_{n}^{i} \right) 
	\label{eq:fcl}
\end{equation}
\begin{equation}
	\mathbf{F}_{c}^{g}=\mathrm{GCA}\left( \mathrm{MiT}\left( \mathbf{F}_{n}^{i} \right) \right) \times \mathrm{MiT}\left( \mathbf{F}_{n}^{i} \right) 
	\label{eq:fcg}
\end{equation}
where LCA and GCA denote Local Channel Attention and Global Channel Attention, respectively, $\mathbf{F}_{c}^{l}$ and $\mathbf{F}_{c}^{g}$ denote locally and globally extracted channel dimension features.

Following Channel Attention, the obtained $\mathbf{F}_{c}^{l}$ and $\mathbf{F}_{c}^{g}$ are sent to Spatial Attention, where they are used to construct global and local attention feature maps $\mathbf{F}_{s}^{g}$,  $\mathbf{F}_{s}^{l}$ in channel dimension. The relevant formulas are as follows:
\begin{equation}
	\mathbf{F}_{s}^{l}=\mathrm{LSA}\left( \mathrm{MiT}\left( \mathbf{F}_{n}^{i} \right) \right) \times \mathbf{F}_{c}^{l}+\mathbf{F}_{c}^{l}
	\label{eq:fsl}
\end{equation}
\begin{equation}
	\mathbf{F}_{s}^{g}=\mathrm{GSA}\left( \mathrm{MiT}\left( \mathbf{F}_{n}^{i} \right) \right) \times \mathbf{F}_{c}^{g}+\mathbf{F}_{c}^{g}
	\label{eq:fsg}
\end{equation}
where LSA and GSA correspond to Local Spatial Attention and Global Spatial Attention in Fig.~\ref{fig:OFAM}. $\mathbf{F}_{s}^{l}$ and $\mathbf{F}_{s}^{g}$ are the two output feature layers of Spatial Attention, which weight local and global information in the spatial dimension, respectively. Unlike Channel Attention, the topic part of Spatial Attention is symmetric, with only Local Spatial Attention and Global Spatial Attention differing.

Following the extraction of global and local information in the channel and spatial dimensions, the features are fused to produce the final output feature map.
\begin{equation}
	\mathbf{F}_{n+1}^{i}=\mathrm{MiT}\left( \mathbf{F}_{n}^{i} \right) \times \mathbf{F}_{s}^{l}\times \mathbf{F}_{s}^{g}+\mathbf{F}_{s}^{l}+\mathbf{F}_{s}^{g}
	\label{eq:outf}
\end{equation}

Each Attention Module in OFAM is detailed in depth in Fig.~\ref{fig:OFAM-detail}. The processing of the feature maps in each section is shown below:
\begin{enumerate}
	\item Part (a) of the diagram depicts a simple convolutional neural network that incorporates the layers of convolution, pooling, and so on by linking them in sequence, as shown in Eq:
	\begin{align}
		\mathbf{F}_{out}=\mathrm{\sigma}\left( \mathrm{Conv}^{3\times 3}\left( \mathrm{LAP}\left( \mathbf{F}_{in} \right) \right) \right) 
	\end{align}
	where $\mathrm{\sigma}$ denotes the Softmax activation function, $\mathrm{LAP}$ denotes Local Channel Attention, $\mathrm{Conv}^{3\times 3}\left( \cdot \right) $ is a convolutional layer with a convolutional kernel size of 3$\times $3, and $\mathbf{F}_{in}$ and $\mathbf{F}_{out}$ denote the input and output, respectively.
	\item Part (b) in Fig.~\ref{fig:OFAM-detail} tends to extract global features compared to part (a) in the design of the pooling layer, and we picked three different sizes of convolution to extract the input features, which are 3$\times$3, 5$\times$5, and 7$\times$7. Part (b) can be written as follows:
	\begin{align}
		\begin{split}
			\mathbf{F}_{out}&=\mathrm{\sigma}\left( \mathrm{Conv}^{3\times 3}\left( \mathrm{Concat}\left( \mathrm{Conv}\left( \mathbf{F}_{in} \right) \right) \right) \right)\\
			\mathrm{Conv}\left( \cdot \right) &=\left\{ \mathrm{Conv}^{3\times 3}\left( \cdot \right) ,\mathrm{Conv}^{5\times 5}\left( \cdot \right) ,\mathrm{Conv}^{7\times 7}\left( \cdot \right) \right\}
		\end{split}
	\end{align}
	where $\mathrm{Concat}$ denotes the splicing of the input feature $\mathbf{F}_{in}$ in the channel dimension after three different convolutions and scaling to a uniform size.
	\item Parts (a) and (b) weight the feature maps solely in the channel dimension, but they do not examine the relationship between the convolution kernel and the input feature maps for different convolution sizes, therefore we devised part (c) to address this issue. This can be stated mathematically as follows:
	\begin{align}
		\begin{split}
			\mathbf{F}_{mid}&=\mathrm{\sigma}\left( \prod_{i=1}^2{\mathrm{ConvG}_{i}\left( \mathrm{Conv}\left( \mathbf{F}_{in} \right) \right)} \right)\\
			\mathbf{F}_{out}&=\mathbf{F}_{in}\times \mathbf{F}_{mid}
		\end{split}
	\end{align}
	where $\mathrm{ConvG}_{i}$ denotes that the features are first subjected to a convolution operation with a convolution kernel size of 3$\times$3, followed by the GeLU activation function~\cite{hendrycks2016gaussian}.
	\item We created the module depicted in (d) to use the ability of the interaction between different convolutional kernels for the extraction of global information, with the goal of weight extraction of global information at the spatial level. The following are the calculating formulas:
	\begin{align}
		\begin{split}
			\mathbf{F}_{mid}&=\gamma \left( \mathrm{Conv}^{5\times5}\left( \mathbf{F}_{in} \right) \times \mathrm{Conv}^{7\times7}\left( \mathbf{F}_{in} \right) \right)\\
			\mathbf{F}_{out}&=\mathrm{ConvS}\left( \mathbf{F}_{mid}\times \mathrm{Conv}^{3\times3}\left( \mathbf{F}_{in} \right) \right)
		\end{split}
	\end{align}
	where $\gamma $ denotes the GeLU activation function, and $\mathrm{Conv}^{5\times5}$ and $\mathrm{Conv}^{7\times7}$ denote convolutional layers with convolutional kernel sizes of 5$\times$5 and 7$\times$7, respectively.
\end{enumerate} 
\begin{figure}[h]
	\centering
	\includegraphics[width=1\linewidth]{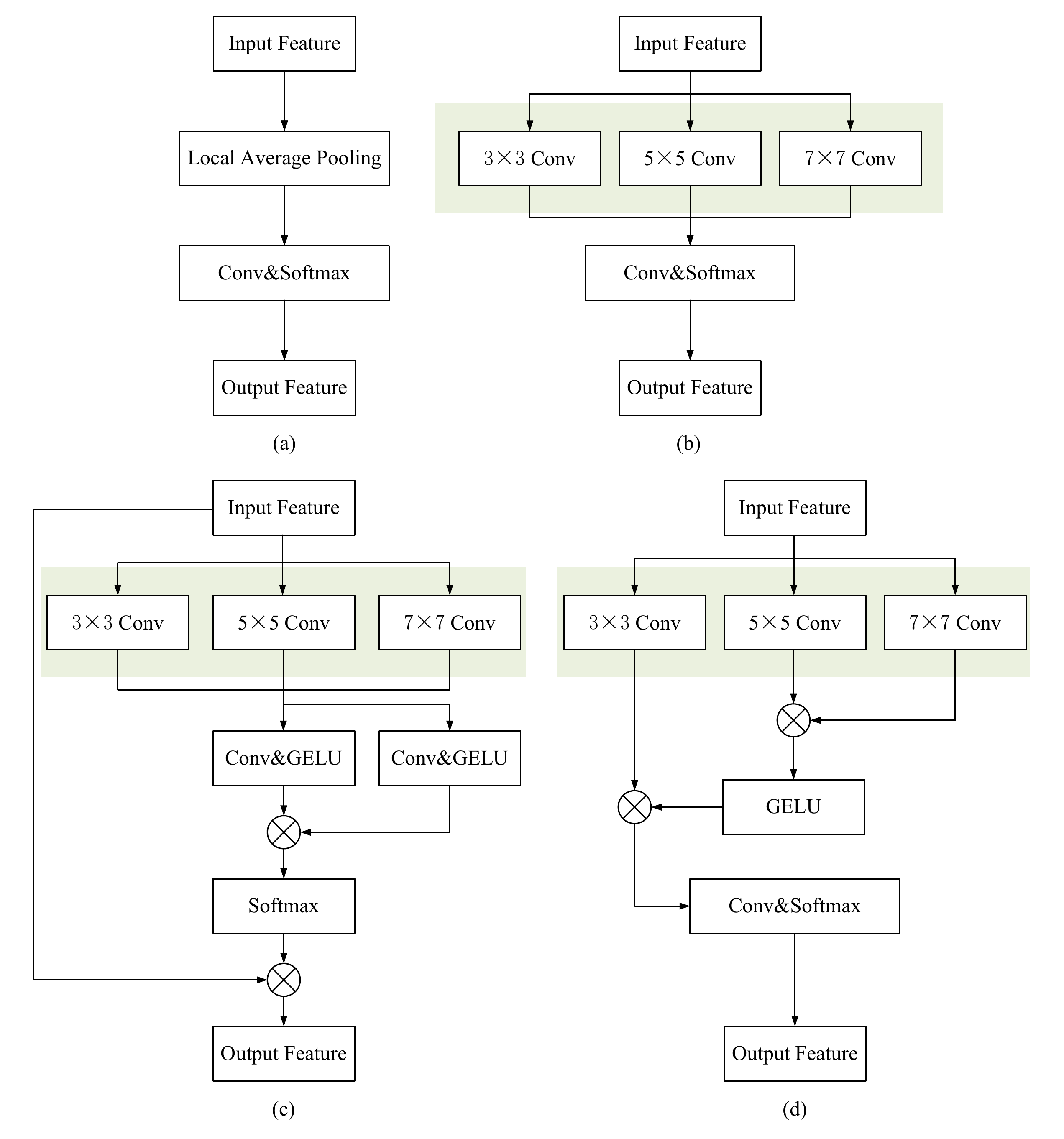}
	\caption{Parts (a), (b), (c), and (d) of the OFAM schematic diagrams depict Local Channel Attention, Global Channel Attention, Local Spatial Attention, and Global Spatial Attention, respectively, in Fig.~\ref{fig:OFAM}.}
	\label{fig:OFAM-detail}
\end{figure}

In the model feature extraction section, we combine the MiT feature extractor with OFAM. The global and local information in the feature map is retrieved simultaneously in both channel and spatial dimensions, thereby aggregating the positional and spectral information in the remote sensing image.

\subsection{FA Module}
\label{sec:FA-Module}
After Backbone, we created a feature aggregation module called FA (Feature Alignment) Module to better aggregate the deep features produced by feature extraction for bi-temporal images. The FA Module construction is depicted in Fig.~\ref{fig:FA-Module}. We integrate MixFFN from SegFormer~\cite{xie2021segformer} after FDAF in ChangerEX~\cite{fang2023changer} to improve feature representation and contextual comprehension when performing feature extraction in image altering regions. The following are the relevant formulas:
\begin{figure*}[ht]
	\centering
	\includegraphics[width=0.8\linewidth]{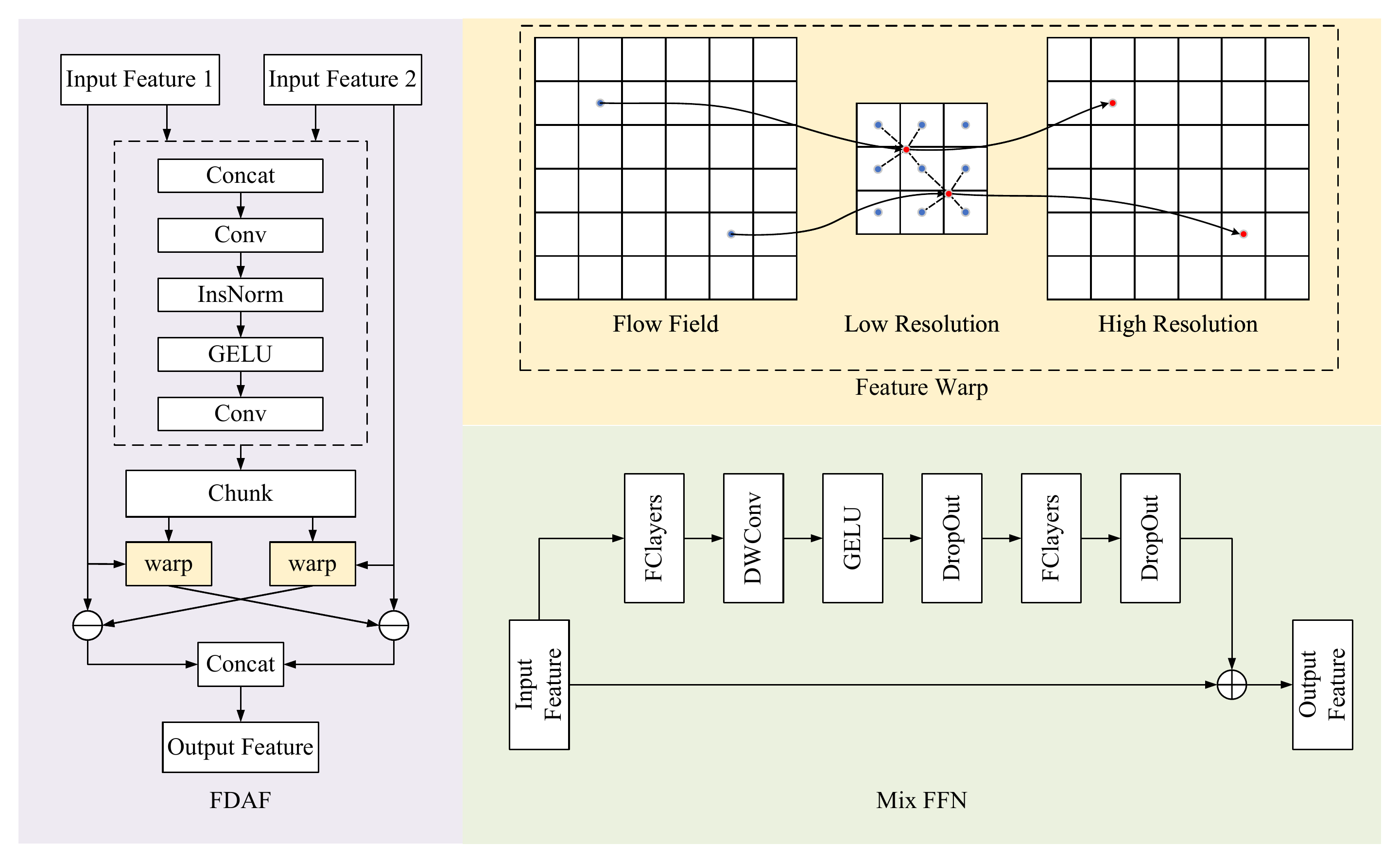}
	\caption{A schematic representation of our FA Module, which is separated into two main portions, FDAF and MixFFN, which are distinguished by various colored backgrounds.}
	\label{fig:FA-Module}
\end{figure*}
\begin{equation}
	\begin{split}
		\mathbf{F}_{con}&=\mathrm{Concat}\left( \mathbf{F}_{in1},\mathbf{F}_{in2} \right)\\
		\mathbf{F}_{flow}&=\mathrm{Conv}\left( \gamma \left( \mathrm{InsNorm}\left( \mathrm{Conv}\left( \mathbf{F}_{con} \right) \right) \right) \right)\\
		\mathbf{F}_{FDAF}&=\mathrm{Concat}\left( \mathbf{F}_{in}-\mathrm{warp}\left( \mathbf{F}_{flow1},\mathbf{F}_{flow2} \right) \right)
	\end{split}
	\label{eq:FDAF}
\end{equation}
where $\mathbf{F}_{in1}$, $\mathbf{F}_{in2}$ denote the feature maps generated by Backbone respectively, $\mathrm{InsNorm}$ denotes the Instance normalization method~\cite{huang2017arbitrary}, $\gamma $ denotes the GeLU activation function, and warp is the Feature Warp in the upper right corner of the Fig.~\ref{fig:FA-Module}.

In FDAF, we first splice the two input features in channel dimension and then insert them into the dashed box on the left side of the figure. Borrowing the idea of flow field in the field of video processing~\cite{tsai2016video}, the authors design a feature alignment method, i.e., warp in Fig.~\ref{fig:FA-Module}, to correct the feature offset problem caused by the dimensional change of the input feature maps after feature extraction is performed.

In warp, the semantic flow field $\Delta _{l-1}$ is generated by bilinear interpolating $\mathbf{F}_{l-1}$ to the same size as $\mathbf{F}_{l}$, then concatenating the two in the channel dimension, and finally a convolutional layer. Following that, using a simple addition operation, each position ${p}_{l-1}$ is mapped to a point ${p}_{l}$ in the preceding layer $l$. Finally, using a bilinear sampling method, the values of the four nearby pixels are linearly interpolated to approximate the FAM's final output $\mathbf{F}_{l}\left( {p}_{l-1} \right) $. The following are the relevant formulas for the aforementioned calculations:
\begin{equation}
	\begin{split}
		\Delta _{l-1}&=\mathrm{Conv}_{l}\left( \mathrm{Concat}\left( \mathbf{F}_{l},\mathbf{F}_{l-1} \right) \right) \\
		{p}_{l}&={p}_{l-1}+\frac{\Delta _{l-1}\left( {p}_{l-1} \right)}{2}\\
		\mathbf{F}_{l}\left( {p}_{l-1} \right) &=\mathbf{F}_{l}\left( {p}_{l} \right) =\sum_{{p}\in N\left( {p}_{l} \right)}{\mathrm{\omega}_{p}\mathbf{F}_{l}\left( p \right)}
	\end{split}
	\label{eq:warp}
\end{equation}
where $N\left( {p}_{l} \right) $ denotes the neighborhood of the deformation point ${p}_{\mathrm{l}}$ in $\mathbf{F}_{l}$ and $\mathrm{\omega}_{p}$ denotes the bilinear kernel weights.

Considering the information interaction between bi-temporal RS images and inspired by ChangerEx, we introduce FDAF into the method of this paper and simultaneously insert MixFFN after FDAF to improve the feature expression ability after information fusion between bi-temporal phases.

\subsection{Feature Decouple Module}
Some of the image change edges in the RSCD datasets were found to be blurred. This is due in part to the long shadows cast by tilt photography on ground buildings, and in part to blurring of image regions of interest caused by image quality issues in remote sensing photographs such as overexposure, thin clouds, and so on.

Meanwhile, in the RSCD datasets, detection accuracy is high relative to the edges of the modified region due to consistent semantic information throughout the building, indicating homogeneity. In order to solve the aforementioned challenges, we expect to decouple the changing region interior and edges throughout the training process, which will allow us to extract the region boundary on the one hand and effectively minimize the computation on the other.

\begin{figure}[htbp]
	\centering
	\includegraphics[width=1\linewidth]{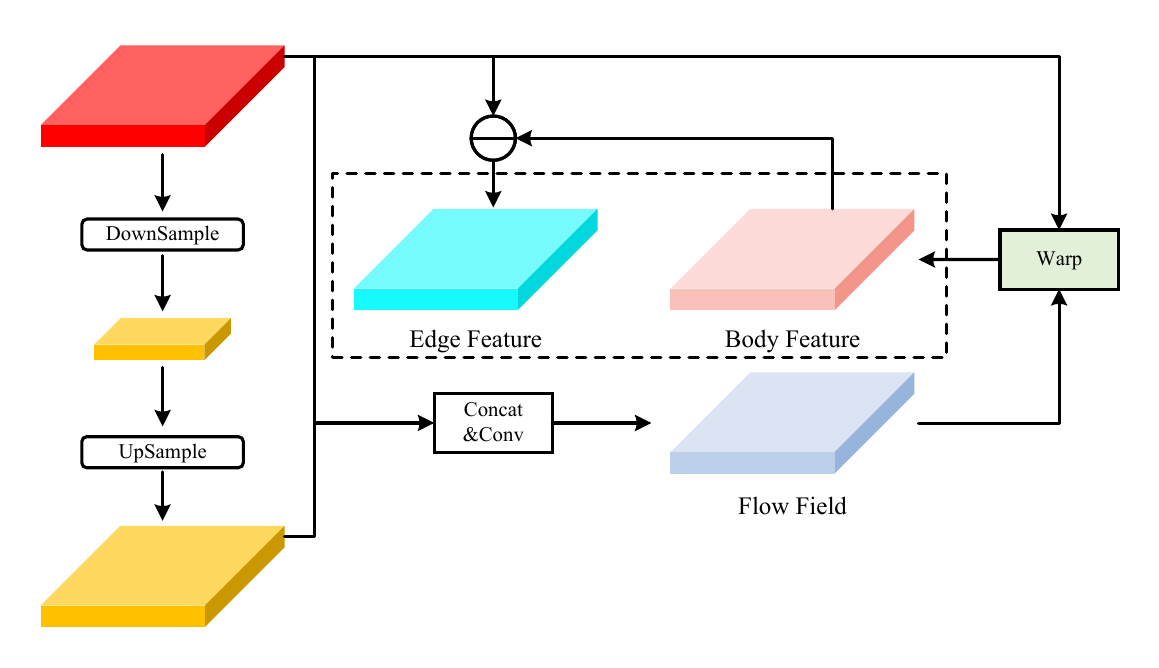}
	\caption{Illustration of our proposed Decouple Module.}
	\label{fig:Decouple-Module}
\end{figure}

As a result, we use the flow field concept and add the Decouple Module after feature decoding in the model to successfully extract the boundary of the changing region throughout the training process. The decouple module is depicted in Fig.~\ref{fig:Decouple-Module}.

We initially sample the input feature map $\mathbf{F}_{in}$ twice (DownSample and UpSample) in Fig.~\ref{fig:Decouple-Module} to boost its semantic information without affecting the feature size. In sectionrefsec:FA-Module, we use Warp to correct the features of $\mathbf{F}_{in}$ to get $\mathbf{F}_{body}$, and then subtract $\mathbf{F}_{in}$ from $\mathbf{F}_{body}$ to produce $\mathbf{F}_{edge}$.The following are the relevant formulas:
\begin{equation}
	\begin{split}
		\mathbf{F}_{flow}&=\mathrm{ConcatConv}\left( \mathbf{F}_{in},\mathrm{DownUp}\left( \mathbf{F}_{in} \right) \right) \\
		\mathbf{F}_{body}&=\mathrm{Warp}\left( \mathbf{F}_{flow},\mathbf{F}_{in} \right) \\
		\mathbf{F}_{edge}&=\mathbf{F}_{in}-\mathbf{F}_{body}
	\end{split}
\end{equation}
where $\mathrm{DownUp}$ indicates that $\mathbf{F}_{in}$ is downsampled before being upsampled.

After passing the features via the Decouple Module during the model training process, the features are successfully classified as edge features and body features. To the best of our knowledge, we are the first in the field of RSCD to do so. This substantially enhances the model's prediction capacity and, to some extent, reduces the number of parameters in the model.

\section{Experimental Results and Analysis}
\label{sec:experiments}
In this section, we first introduce the dataset, experimental environment, and validation metrics used in this paper's experiments, then compare the model of this paper to other models, conduct ablation experiments to evaluate the effect of each module, and finally visualize some of the feature maps generated during the model's training process.
\subsection{Experimental Setup}
For this experiment, the three public RSCD datasets listed below were employed.

\textbf{DSIFN-CD}~\cite{zhang2020deeply} is derived from six Chinese cities, including Beijing, and was manually collected in Google Earth. It is a publicly available binary change detection dataset with a spatial resolution of 2m that includes changes to roads, buildings, agriculture, and water bodies. During the experimental process, we cropped each image to 512$\times$512, and the test set in the original dataset was of lower quality, so we divided the original training set into a training set and a validation set, and we used the original validation set as the test set, and the dataset now has 3000/600/340 training/validation/test respectively.

\textbf{S2Looking}~\cite{shen2021s2looking} is a publicly available dataset of 5000 pairs of bi-temporal RS images broken into 3500/500/1000 training/validation/test sets with a spatial resolution of 0.5~0.8m and a size of 1024$\times$1024 for each image.

\textbf{WHU-CD}~\cite{ji2018fully} is a publicly available CD dataset of RS image that covers the area of Christchurch, New Zealand that was struck by a magnitude 6.3 earthquake in February 2011 and rebuilt in subsequent years. The dataset consists of aerial imagery acquired in April 2012 and contains 12,796 buildings in 20.5 square kilometers (16,077 buildings in the same area in the 2016 dataset). The original size of the dataset was 32507$\times$15345 with a resolution of 0.075 m, and was cropped to 256$\times$256 for the experiments and the dataset now has 5947/743/744 training/validation/test respectively. The conditions of this dataset such as illumination are desirable, so it is used for the validation of the generalizability of our model.

\begin{figure}[htbp]
	\centering
	\includegraphics[width=1\linewidth]{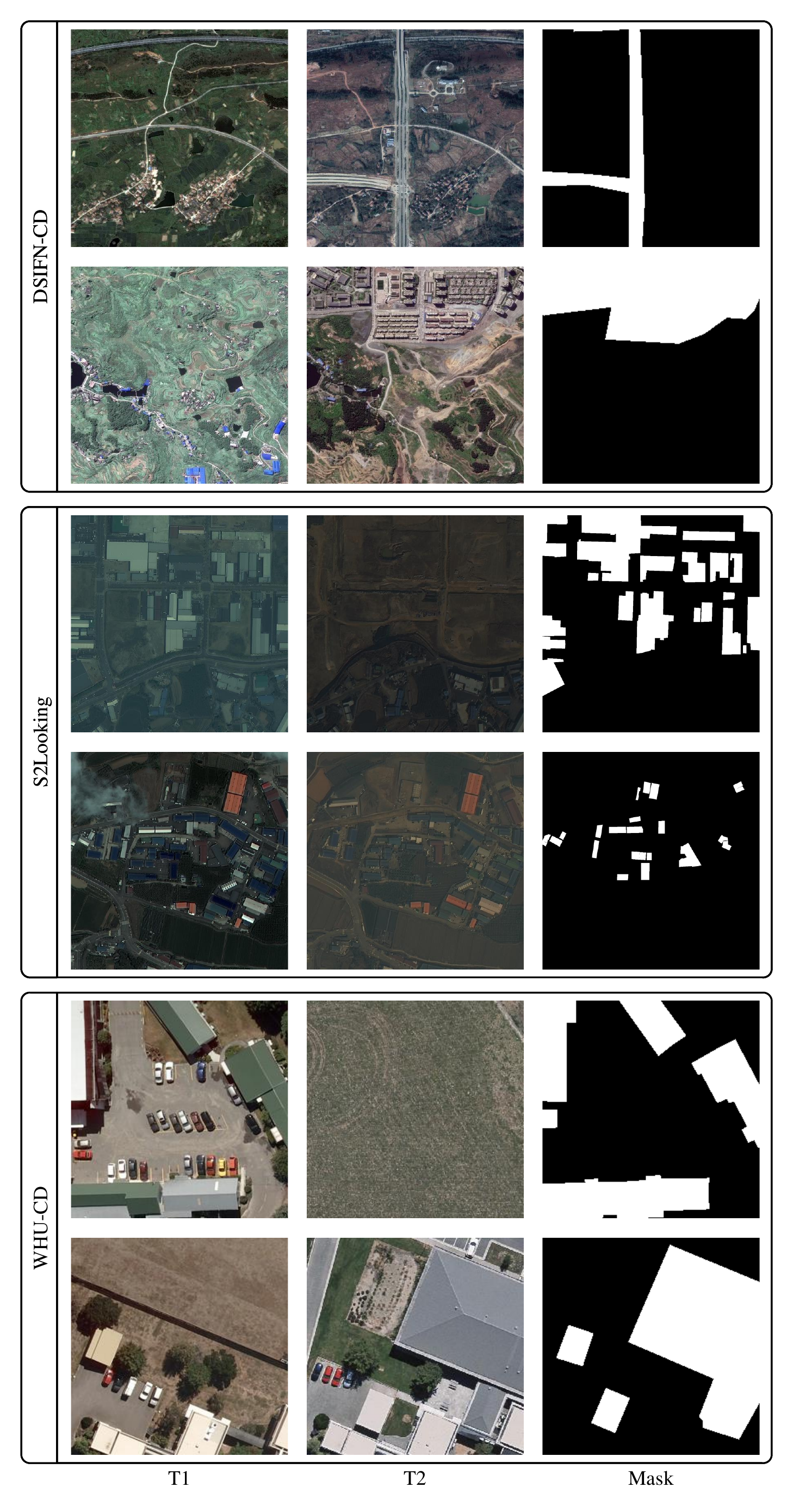}
	\caption{Some of the images in DSIFN-CD, S2Looking and WHU-CD.}
	\label{fig:Datasets}
\end{figure}

Some of the images in DSIFN-CD, S2Looking and WHU-CD are shown in Fig.~\ref{fig:Datasets}. The three columns in the figure are pre-change image, post-change image, and change Mask, respectively.

\subsection{Implementation Details}
This experiment was deployed under PyTorch 2.0.1 and Python 3.8.13. For hardware, we used Intel Xeon E5-2678 v3 @2.50GHz$\times$2, 32GB of RAM as well as used  an NVIDIA RTX 4090 GPU. And for hyper-parameters, we used BCE Loss as the paper's loss function for our experiments and use AdamW as the optimizer, which is formally defined as:
\begin{equation} 
	\begin{split}
		\mathcal{L}_{BCE}&=-\frac{1}{H\times W}\sum_{h=1,w=1}^{H,W} \bigg [{Y\left( h,w \right)}
		\\
		&+\left( 1-Y\left( h,w \right) \right) \cdot \log \left( 1-\hat{Y}\left( h,w \right) \right) \bigg ] 
	\end{split}
	\label{eq:BCELoss}
\end{equation}

\begin{equation}
	\begin{split}
		\theta _{t+1}=\theta _t-\frac{\alpha}{\sqrt{\hat{\upsilon}_t}+\varepsilon}\hat{m}_t-\alpha \lambda \theta _t
	\end{split}
	\label{eq:AdamW}
\end{equation}
in \ref{eq:BCELoss}, where $H\times W$ is the size of the image to be predicted, $Y(h,w)$ is the predicted value of the point $(h,w)$ in the image and $\hat{Y}(h,w)$ is the true value of the point; in \ref{eq:AdamW}, $\theta_{t}$ and $\theta_{t+1}$ denote the parameter values at time steps $t$ and $t+1$, respectively, $\alpha$ is the learning rate, $\hat{m}_t$ and $\hat{\upsilon}_t$ are the exponential moving averages of the first-order and second-order moments, respectively, and $\varepsilon$ is a very small value.

In this paper, we use the Open-CD development kit \cite{fang2023changer} based on OpenMMLab \cite{mmcv} in order to compare the training results of different models in the same experimental environment.

\textbf{Evaluation Metrics}. We used the following metrics for validation to validate the training effect of our proposed BD-MSA: F1-score (F1), Precision (Prec.), Recall (Rec.), and IoU, which are defined as follows:

\begin{equation}\label{eq:Pre}
	\rm Pre.=\frac{TP}{TP+FP}
\end{equation}
\begin{equation}\label{eq:Rec}
	\rm Rec.=\frac{TP}{TP+FN}
\end{equation}
\begin{equation}\label{eq:F1}
	\rm F1=2\times \frac{Pre\times Rec}{Pre+Rec}
\end{equation}
\begin{equation}\label{eq:IoU}
	\rm IoU=\frac{TP}{TP+FP+FN}
\end{equation}
where TP, FP, and FN represents the number of true positive, false positive, and false negative pixels, respectively.

\subsection{Comparison With SOTA Methods}
We compared the approaches mentioned in this work to some SOTA methods, which are listed below:
\begin{itemize}
	\item \textbf{FC-EF, FC-Siam-Di and FC-Siam-Conc}~\cite{daudt2018fully} are built on fully convolutional networks~\cite{long2015fully} with a model structure similar to that of U-Net~\cite{ronneberger2015u}, and they use distinct methodologies to analyze paired image data.
	\item \textbf{BIT}~\cite{chen2021remote} introduces Transformer~\cite{vaswani2017attention} to classic CNN change detection networks, which can more effectively capture long-distance interdependence and complex spatial dynamics.
	\item \textbf{ChangeFormer}~\cite{bandara2022transformer}, unlike typical fully convolutional network-based techniques, ChangeFormer combines a hierarchically structured Transformer encoder and a multilayer perceptron decoder to efficiently capture long range information at multi-scale, enhancing change detection accuracy.
	\item \textbf{ChangerEx-MiT}~\cite{fang2023changer} emphasizes the significance of feature interaction and presents simple but effective interaction mechanisms---AD and feature ``exchange''.
	\item \textbf{HANet}~\cite{han2023hanet} addresses the challenge of data imbalance between changed and unchanged pixels in the change detection task by proposing a stepwise foreground-balanced sampling strategy to improve model learning for changed pixels and employing a concatenated network structure with hierarchical attention to integrate multi-scale features for finer detection.
	\item \textbf{IFNet}~\cite{zhang2020deeply} collects deep features using a fully convolved two-stream architecture and then uses a difference discrimination network and an attention module to identify changes, highlighting the significance of deep supervision in improving border integrity and object internal compactness.
	\item \textbf{SNUNet}~\cite{fang2021snunet}, through tight hopping connections between the encoder and decoder as well as between decoders, SNUNet is able to maintain high-resolution fine-grained features while mitigating pixel uncertainty at the borders of changing targets and deterministic missingness of small targets.
	\item \textbf{STANet}~\cite{chen2020spatial} captures spatial-temporal correlations via a self-attentive method in order to generate more discriminative features. It was divided into three variants, STANet-Base, STANet-Bam, and STANet-Pam.
	\item \textbf{TINY-CD}~\cite{codegoni2023tinycd} employs the Siamese U-Net architecture and a new feature mixing method to optimally utilize low-level information for spatial and temporal domains, while also offering a new spatial-semantic attention mechanism via its Mix and Attention Mask Block (MAMB).
\end{itemize}

\subsection{Main Results}
On the DSIFN-CD and S2Looking datasets, we compared the outcomes of our proposed BD-MSA with previous SOTA approaches in table~\ref{tab:main result}. The \textcolor{red}{top}, \textcolor{blue}{second best}, and \textbf{third best} performers in each evaluation metric are shown in red, blue, and bolded black, respectively. The results reveal that our proposed BD-MSA outperforms the second-best model ChangerEx-MiT on the DSIFN-CD dataset, with an F1 score, IoU of 83.98\% and 72.38\%, respectively, which is 3.11\% and 4.49\% higher. Our suggested BD-MSA achieves an F1 score, IoU of 64.08\% and 47.17\% on the S2Looking dataset, which is 2.1\% and 2.23\% higher than the second-best model IFNet. The results demonstrate that our proposed BD-MSA performs well in the field of RSCD. In the column of \#Param (M), we can see that our Proposed BD-MSA has a modest number of parameters, which is 3.465M; while this indication is not the smallest, it is a comparatively small number of parameters compared to many other techniques.

We also performed the same experiments on WHU-CD to confirm the proposed model's generalizability on other datasets. The findings indicate that, BD-MSA, like DSIFN-CD and S2Looking, achieves the highest metrics of both F1 and IoU on the WHU-CD test set, with an enhancement of 1.16\% and 2.01\%, respectively, over the second-best model. The aforementioned findings demonstrate BD-MSA's superior generalization capability.
\begin{table*}[htbp]
	\caption{Comparison of our proposed BD-MSA with other SOTA methods on DSIFN-CD, S2Looking and WHU-CD datasets. We use different colors to indicate: \textcolor{red}{best}, \textcolor{blue}{second best}, and \textbf{third best}.}
	\label{tab:main result}
	\centering
	\scalebox{0.9}{
	\begin{tabular}{l|c|c|c|cccc|cccc|cccc}
		\toprule
		\midrule
		\multirow{2}{*}{Method} & \multirow{2}{*}{Backbone} & \multirow{2}{*}{\#Param (M)} & \multirow{2}{*}{FLOPs(G)} & \multicolumn{4}{c|}{DSIFN-CD} & \multicolumn{4}{c|}{S2Looking} & \multicolumn{4}{c}{WHU-CD}\\
		&  &  &  & F1 & Prec. & Rec. & IoU & F1 & Prec. & Rec. & IoU & F1 & Prec. & Rec. & IoU\\
		\midrule
		FC-EF \cite{daudt2018fully} & - & 1.353 & 12.976 & 63.44 & 75.97 & 54.47 & 46.46 & 7.65 & \textcolor{blue}{81.36} & 8.95 & 8.77 & 69.73 & 80.32 & 61.6 & 53.52 \\
		FC-Siam-Di \cite{daudt2018fully} & - & 1.352 & 17.54 & 63.41 & 73.23 & 55.92 & 46.43 & 13.19 & \textcolor{red}{83.29} & 15.76 & 15.28 & 64.42 & 55.87 & 76.07 & 47.51 \\
		FC-Siam-Conc \cite{daudt2018fully} & - & 1.548 & 19.956 & 67.68 & 66.83 & 68.56 & 51.15 & 13.54 & 68.27 & 18.52 & 17.05 & 59.62 & 46.88 & 81.9 & 42.47 \\
		BIT \cite{chen2021remote} & ResNet18 & 2.99 & 34.996 & 71.04 & 77.22 & 65.78 & 55.09 & 44.51 & 67.41 & 33.23 & 28.63 & 91.08 & 91.55 & 90.61 & 83.62 \\
		ChangeFormer \cite{bandara2022transformer} & MiT-b1 & 3.847 & 11.38 & \textbf{80.23} & 84.4 & \textcolor{blue}{76.46} & \textbf{66.99} & \textbf{60.92} & \textbf{75.79} & 50.93 & 43.8 & \textcolor{blue}{92.25} & \textcolor{red}{95.39} & 89.31 & \textcolor{blue}{85.62} \\
		ChangerEx-MiT \cite{fang2023changer} & MiT-b0 & 3.457 & 8.523 & \textcolor{blue}{80.87} & \textcolor{blue}{87.93} & 74.87 & \textcolor{blue}{67.89} & 60.01 & 67.52 & 54.0 & \textbf{42.87} & 89.04 & 90.92 & 87.24 & 80.25 \\
		HANet \cite{han2023hanet} & - & 3.028 & 97.548 & 75.91 & 75.9 & \textbf{75.92} & 61.18 & 43.67 & 44.89 & 42.51 & 27.93 & 79.18 & 86.65 & 72.9 & 65.54 \\
		IFNet \cite{zhang2020deeply} & VGG-16 & 35.995 & 323.584 & 79.21 & \textbf{85.54} & 73.75 & 65.58 & \textcolor{blue}{61.98} & 64.96 & 59.27 & \textcolor{blue}{44.91} & 91.09 & 90.74 & 91.45 & 83.64 \\
		SNUNet \cite{fang2021snunet} & - & 3.012 & 46.921 & 76.08 & 78.26 & 74.02 & 61.4 & 48.25 & 60.8 & 39.99 & 31.79 & 74.1 & 66.51 & 83.64 & 58.86 \\
		STANet-Base \cite{chen2020spatial} & ResNet18 & 12.764 & 70.311 & 66.28 & 76.07 & 58.71 & 49.56 & 26.92 & 15.87 & \textcolor{blue}{88.54} & 15.55 & 65.35 & 50.43 & \textbf{92.8} & 48.53 \\
		STANet-Bam \cite{chen2020spatial} & ResNet18 & 12.846 & 391.168 & 61.48 & 70.83 & 54.31 & 44.39 & 27.27 & 16.11 & \textcolor{red}{88.68} & 15.79 & 67.88 & 52.47 & \textcolor{blue}{96.12} & 51.38 \\
		STANet-Pam \cite{chen2020spatial} & ResNet18 & 13.356 & 512 & 37.84 & 76.13 & 25.18 & 23.34 & 23.73 & 13.85 & \textbf{82.79} & 13.46 & 76.64 & 63.64 & \textcolor{red}{96.31} & 62.13 \\
		TINY-CD \cite{codegoni2023tinycd} & EfficientNet & 0.285 & 5.791 & 74.71 & 76.37 & 73.12 & 59.63 & 54.5 & 63.81 & 47.56 & 37.46 & \textbf{91.17} & \textbf{92.15} & 90.21 & \textbf{83.77} \\
		\midrule
		\textbf{BD-MSA (Ours)} & MiT-b0 & 3.465 & 12.658 & \textcolor{red}{83.98} & \textcolor{red}{88.01} & \textcolor{red}{80.3} & \textcolor{red}{72.38} & \textcolor{red}{64.08} & 70.44 & 58.77 & \textcolor{red}{47.14} & \textcolor{red}{93.41} & \textcolor{blue}{94.3} & 92.53 & \textcolor{red}{87.63} \\
		\midrule
		\bottomrule
	\end{tabular}
	}
\end{table*}

\begin{figure*}
	\centering
	\includegraphics[width=1\linewidth]{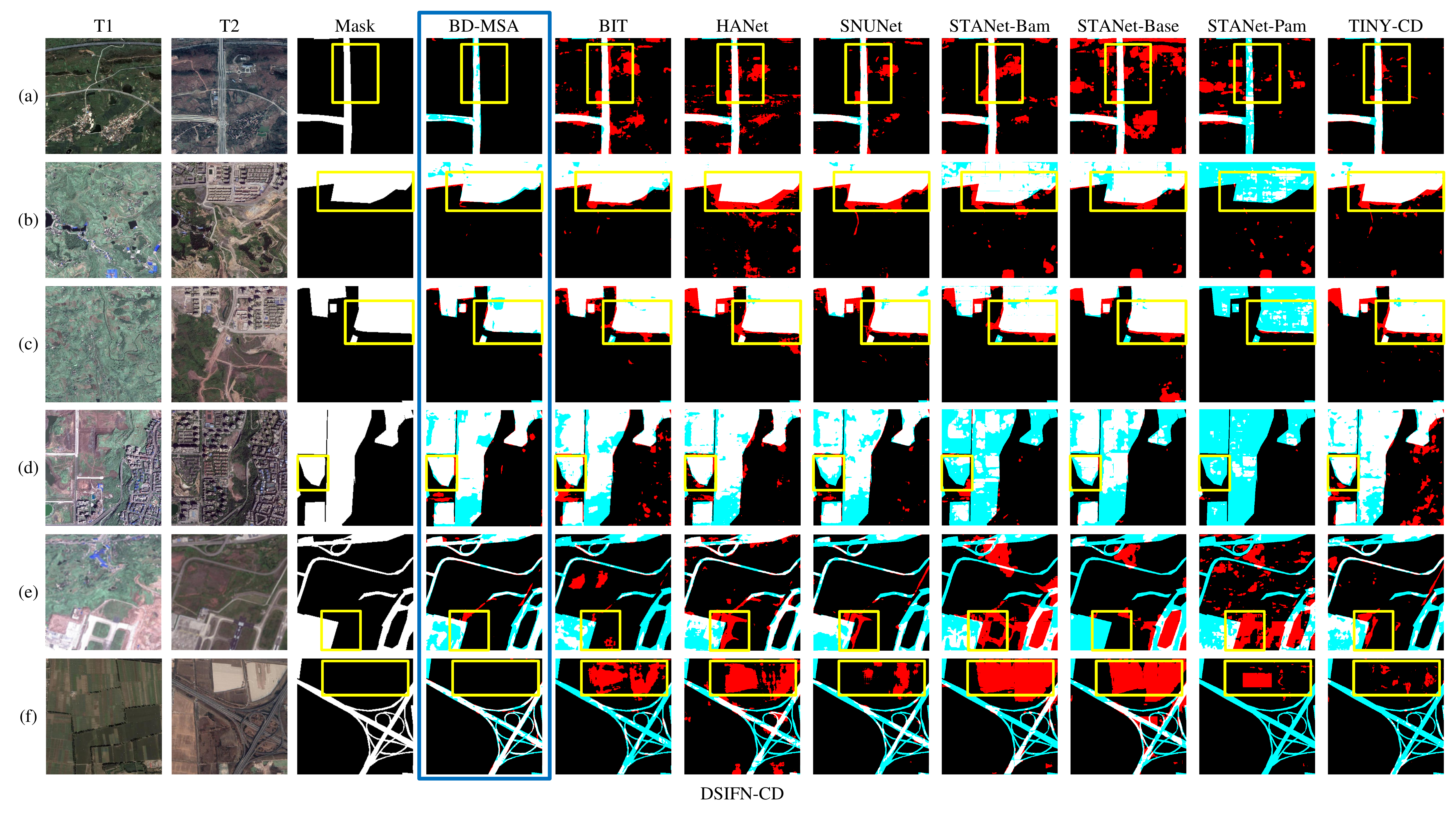}
	\caption{Comparative experimental visualization results for each model on the DSIFN-CD test sets, which different colored regions denote \textcolor[RGB]{255, 0, 0}{FP}, \textcolor[RGB]{0, 255, 255}{FN}, and \textbf{TN}, respectively, and where the white region is TP.}
	\label{fig:main_results_dsifn}
\end{figure*}

\begin{figure*}
	\centering
	\includegraphics[width=1\linewidth]{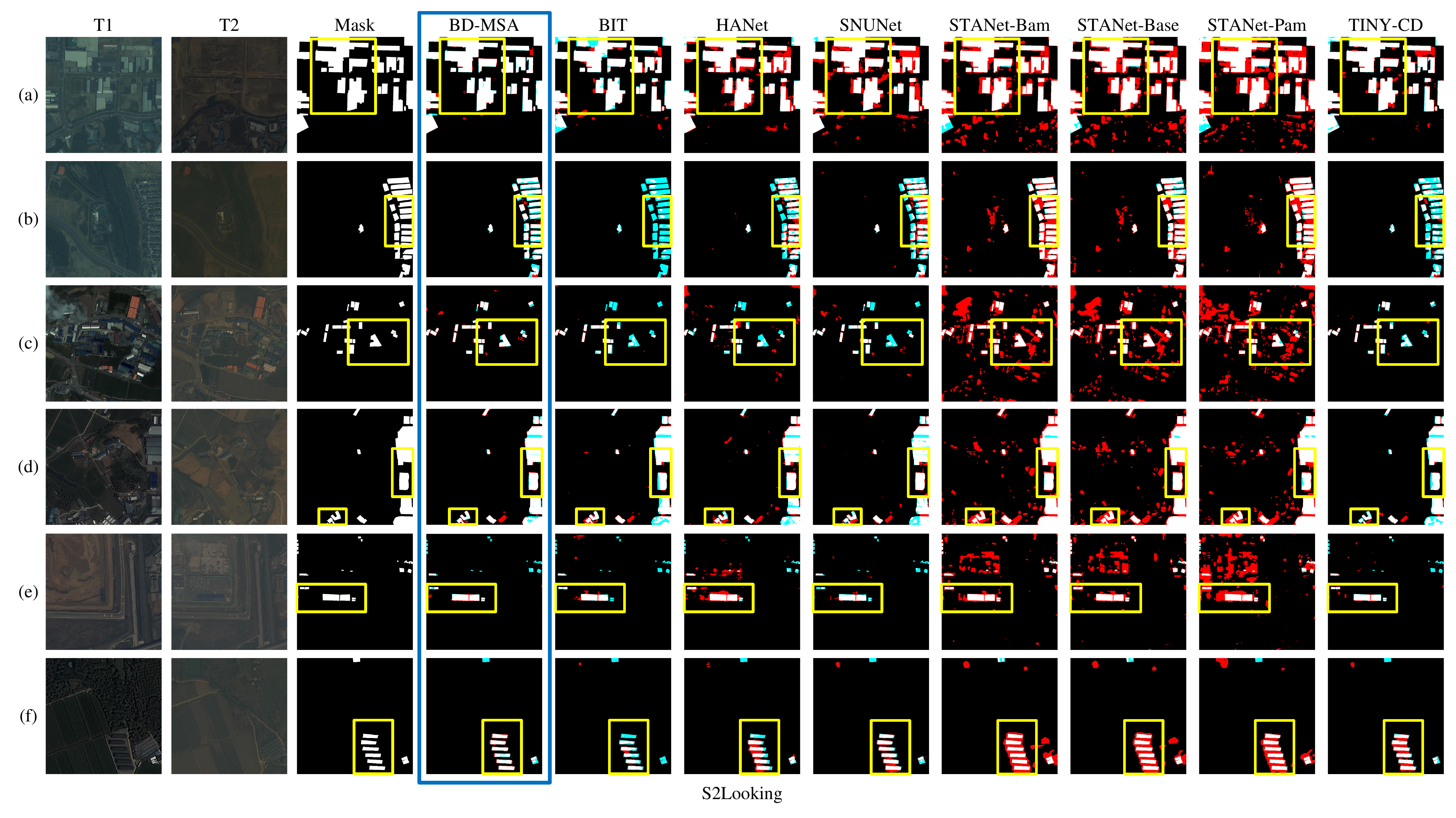}
	\caption{Comparative experimental visualization results for each model on the S2Looking test sets, which different colored regions denote \textcolor[RGB]{255, 0, 0}{FP}, \textcolor[RGB]{0, 255, 255}{FN}, and \textbf{TN}, respectively, and where the white region is TP.}
	\label{fig:main_results_s2looking}
\end{figure*}

We visualized the prediction results in the DSIFN-CD and S2Looking datasets to compared the method of this research with other methods in prediction results, as shown in Fig.~\ref{fig:main_results_dsifn} and Fig.~\ref{fig:main_results_s2looking}. varied hues in the graphic represent the model's varied prediction results for each pixel during the prediction phase. Simply said, the greater the proportion of white and black patches in the figure to the total image, the better the model's prediction outcome. 

\begin{figure*}
	\centering
	\includegraphics[width=1\linewidth]{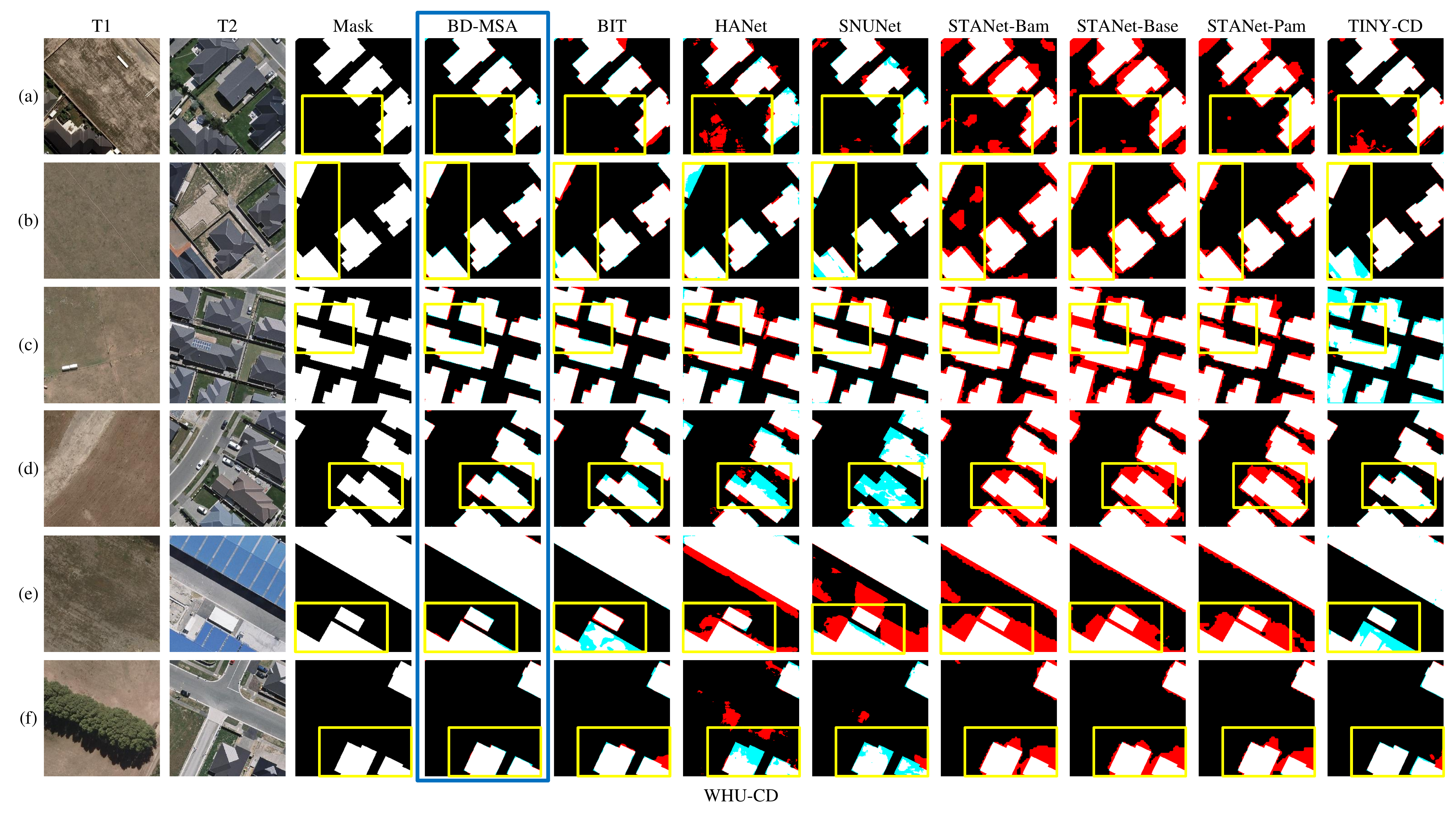}
	\caption{Comparative experimental visualization results for each model on the WHU-CD test sets, which different colored regions denote \textcolor[RGB]{255, 0, 0}{FP}, \textcolor[RGB]{0, 255, 255}{FN}, and \textbf{TN}, respectively, and where the white region is TP.}
	\label{fig:main_results_whu}
\end{figure*}

We specifically chose six photographs at random from each of DSIFN-CD and S2Looking as a test, and it is evident that the method in this work outperforms the other methods in terms of prediction outcomes. In Fig.~\ref{fig:main_results_dsifn}(a), (e), and (f), our proposed approach effectively mitigates misclassification for non-changing regions when making predictions; see Fig.~\ref{fig:main_results_dsifn}, other models' predictions for the boundary of the changing regions are generally confusing in (b), (c), and (d), however the model in this research solves the problem to a degree. Although certain models, such as STANet-Pam, have fewer mispredictions within the limits of the change region, they have a high missed detection rate, implying that the model cannot identify the boundaries well. The improvement of this paper's model over other models, for S2Looking, is mostly in the precision of modifying the region's boundary and the effective decrease of the adhesion phenomenon between buildings. Refer to Fig.~\ref{fig:main_results_s2looking}, in (d), (e), BD-MSA predicts the edges of changing zones more accurately; in (a), (a), and (f), BD-MSA successfully mitigates the adhesion phenomena between buildings with more compact layouts.

To confirm the generalizability of BD-MSA, we conducted tests akin to the ones described above, randomly selecting six images from the WHU-CD test set to test each model. The experimental outcomes are displayed in Fig.~\ref{fig:main_results_whu}. In comparison to the other models, BD-MSA demonstrates satisfying test results that are extremely near to the mask in many images, similar to the findings on the DSIFN-CD and S2Looking test sets.

\begin{figure*}[h]
	\centering
	\includegraphics[width=1\linewidth]{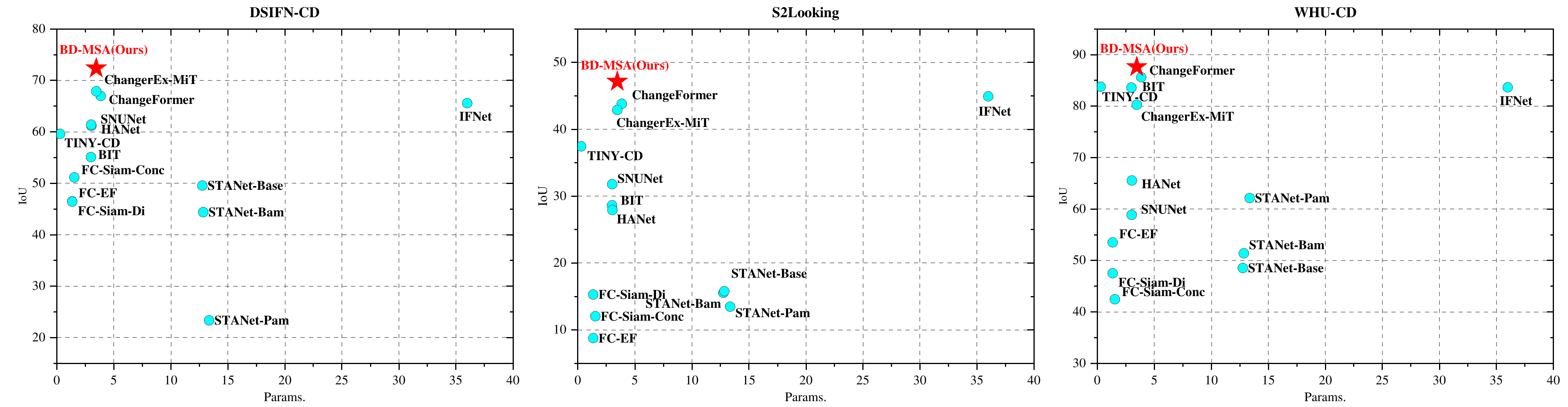}
	\caption{Params. and IoU of different models on the two datasets, the top and bottom parts of the figure show the evaluation results of each model on DSIFN-CD, S2Looking and WHU-CD, respectively.}
	\label{fig:Params-IoU}
\end{figure*}

To compared IoU as well as Params. between multiple models at the same time, we plotted the color mapping for the test results of different models on both datasets, as shown in Fig.~\ref{fig:Params-IoU}. Each point in the graphic represents a model, with the horizontal axis representing the model's parameters and the vertical axis representing the IoU of each model on the three datasets. The closer the model is to the upper left corner of the figure, the higher the accuracy detection, while the model takes less arithmetic power. Our proposed BD-MSA may be seen in the upper left corner, suggesting that the IoU reaches its maximum value and the number of parameters is lower than in most models.

Furthermore, the preceding conclusions show that the model in this study migrates better across devices than alternative models, particularly for machines with weaker arithmetic capability.

\subsection{Ablation Studies}
We conduct ablation tests on OFAM, MixFFN, and the Decouple Module, respectively, to validate the influence of different modules on our proposed model. 

The nomenclature of the models in the ablation experiments is as follows:
\begin{itemize}
	\item \textbf{Baseline}: MiT + FDAF + Predict layer.
	\item \textbf{BD-MSA-1-1}: Baseline + MixFFN.
	\item \textbf{BD-MSA-1-2}: Baseline + Decouple.
	\item \textbf{BD-MSA-1-3}: Baseline + OFAM.
	\item \textbf{BD-MSA-2-1}: Baseline + MixFFN + Decouple Module.
	\item \textbf{BD-MSA-2-2}: Baseline + MixFFN + OFAM.
	\item \textbf{BD-MSA-2-3}: Baseline + Decouple Module + OFAM.
	\item \textbf{BD-MSA}: Baseline + MixFFN + Decouple Module + OFAM.
\end{itemize}

The results of each ablation experiments are shown in Table~\ref{tab:ablation_dsifn} ,\ref{tab:ablation_s2looking} and~\ref{tab:ablation_whu}.
\begin{table}[ht]
	\caption{Results of ablation experiments on DSIFN-CD Test. We use different colors to indicate: \textcolor{red}{best}, \textcolor{blue}{second best}, and \textbf{third best}.}
	\label{tab:ablation_dsifn}
	\centering
	\scalebox{0.8}{
	\begin{tabular}{c|ccc|cccc}
		\toprule
		\midrule
		Module & +MixFFN & +Decouple & +OFAM & F1 & Prec. & Rec. & IoU \\
		\midrule
		Baseline &  &  &  & 80.64 & 84.73 & 76.92 & 67.56 \\
		BD-MSA-1-1 & \checkmark &  &  & 80.87 & \textbf{87.93} & 74.87 & 67.89 \\
		BD-MSA-1-2 &  & \checkmark &  & 80.87 & \textcolor{blue}{87.94} & 74.85 & 67.89 \\
		BD-MSA-1-3 &  &  & \checkmark & 81.7 & 86.64 & 77.3 & 69.06 \\
		BD-MSA-2-1 & \checkmark & \checkmark &  & 81.04 & 86.19 & 76.46 & 68.12 \\
		BD-MSA-2-2 & \checkmark &  & \checkmark & \textcolor{blue}{82.82} & 86.92 & \textbf{79.1} & \textcolor{blue}{70.68} \\
		BD-MSA-2-3 &  & \checkmark & \checkmark & \textbf{82.77} & 86.29 & \textcolor{blue}{79.52} & \textbf{70.6} \\
		BD-MSA & \checkmark & \checkmark & \checkmark & \textcolor{red}{83.98} & \textcolor{red}{88.01} & \textcolor{red}{80.3} & \textcolor{red}{72.38} \\
		\midrule
		\bottomrule
	\end{tabular}
	}
\end{table}

This show that adding each module improves the assessment metrics F1 and IoU when compared to the baseline, with F1 being able to synthesize Prec. and Rec. When only one module is added, adding OFAM results in the greatest improvement in assessment metrics, which we assume is related to the fact that OFAM is added to all four phases of the backbone.

\begin{table}[htbp]
	\caption{Results of ablation experiments on S2Looking Test. We use different colors to indicate: \textcolor{red}{best}, \textcolor{blue}{second best}, and \textbf{third best}.}
	\label{tab:ablation_s2looking}
	\centering
	\scalebox{0.8}{
	\begin{tabular}{c|ccc|cccc}
		\toprule
		\midrule
		Module & +MixFFN & +Decouple & +OFAM & F1 & Prec. & Rec. & IoU \\
		\midrule
		Baseline &  &  &  & 56.72 & 70.21 & 47.58 & 39.59 \\
		BD-MSA-1-1 & \checkmark &  &  & 60.01 & 67.52 & 54.0 & 42.87 \\
		BD-MSA-1-2 &  & \checkmark &  & 59.73 & 70.33 & 51.9 & 42.58 \\
		BD-MSA-1-3 &  &  & \checkmark & 61.94 & 70.38 & 55.31 & 45.93 \\
		BD-MSA-2-1 & \checkmark & \checkmark &  & 63.27 & \textcolor{red}{75.36} & 54.52 & 46.27 \\
		BD-MSA-2-2 & \checkmark &  & \checkmark & \textcolor{blue}{63.31} & \textcolor{blue}{73.49} & \textbf{55.61} & \textcolor{blue}{46.32} \\
		BD-MSA-2-3 &  & \checkmark & \checkmark & \textbf{63.29} & \textbf{71.23} & \textcolor{blue}{56.94} & \textbf{46.28} \\
		BD-MSA & \checkmark & \checkmark & \checkmark & \textcolor{red}{64.08} & 70.44 & \textcolor{red}{58.77} & \textcolor{red}{47.14} \\
		\midrule
		\bottomrule
	\end{tabular}
	}
\end{table}

\begin{table}[htbp]
	\caption{Results of ablation experiments on WHU-CD Test. We use different colors to indicate: \textcolor{red}{best}, \textcolor{blue}{second best}, and \textbf{third best}.}
	\label{tab:ablation_whu}
	\centering
	\scalebox{0.8}{
	\begin{tabular}{c|ccc|cccc}
		\toprule
		\midrule
		Module & +MixFFN & +Decouple & +OFAM & F1 & Prec. & Rec. & IoU \\
		\midrule
		Baseline &  &  &  & 89.04 & 90.92 & 87.24 & 80.25 \\
		BD-MSA-1-1 & \checkmark &  &  & 89.96 & 90.63 & 89.3 & 81.75 \\
		BD-MSA-1-2 &  & \checkmark &  & 89.32 & 93.73 & 85.31 & 80.7 \\
		BD-MSA-1-3 &  &  & \checkmark & 91.02 & 94.14 & 88.11 & 83.52 \\
		BD-MSA-2-1 & \checkmark & \checkmark &  & 92.25 & \textcolor{red}{95.39} & 89.31 & 85.62 \\
		BD-MSA-2-2 & \checkmark &  & \checkmark & \textbf{92.48} & \textcolor{blue}{94.41} & \textbf{90.63} & \textbf{86.01} \\
		BD-MSA-2-3 &  & \checkmark & \checkmark & \textcolor{blue}{92.59} & \textbf{94.33} & \textcolor{blue}{90.9} & \textcolor{blue}{86.2} \\
		BD-MSA & \checkmark & \checkmark & \checkmark & \textcolor{red}{93.41} & 94.3 & \textcolor{red}{92.53} & \textcolor{red}{87.63} \\
		\midrule
		\bottomrule
	\end{tabular}
	}
\end{table}

\begin{figure*}[ht]
	\centering
	\includegraphics[width=1\linewidth]{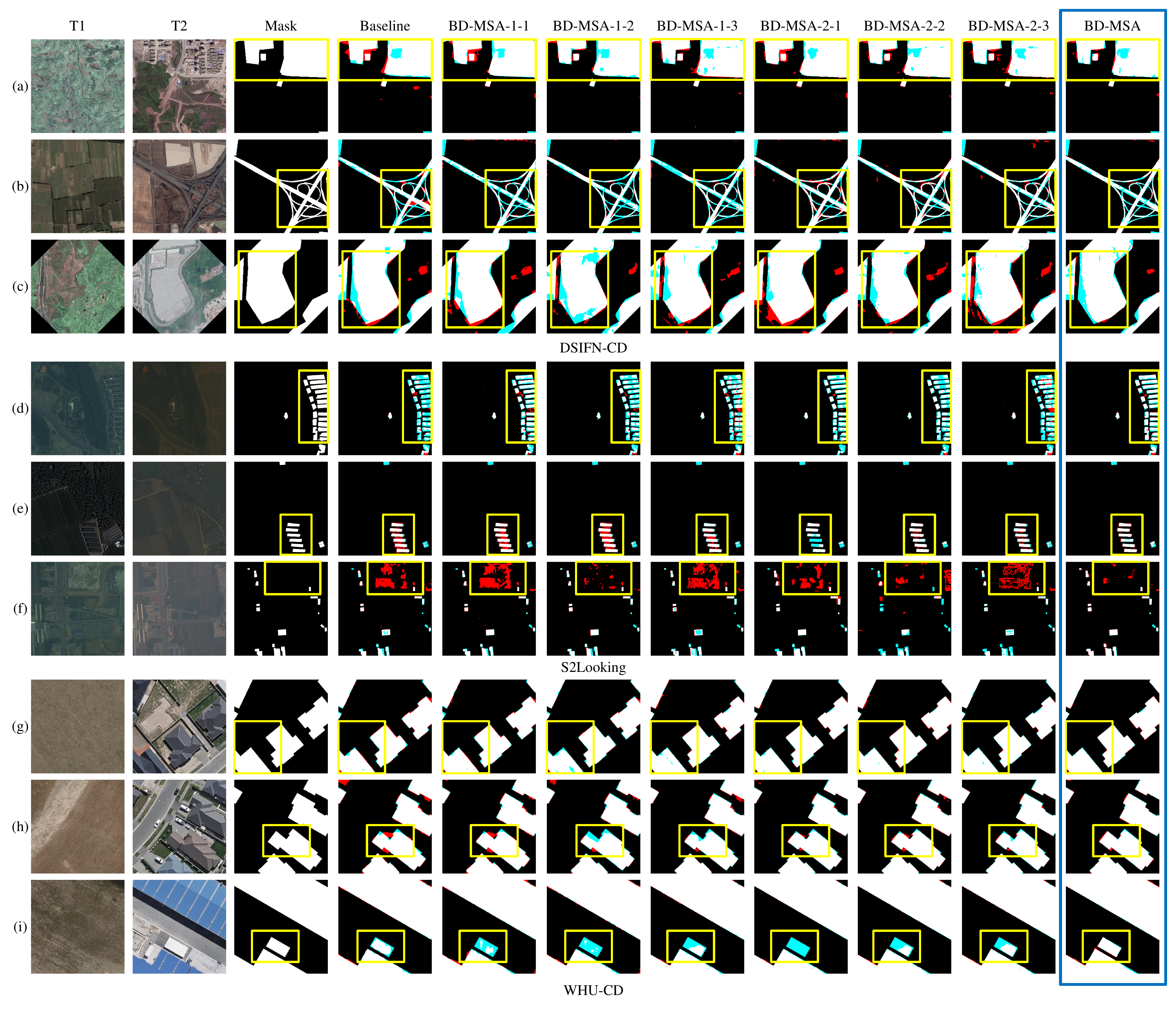}
	\caption{The results of ablation experiments for each model on the DSIFN-CD, S2Looking and WHU-CD test sets.}
	\label{fig:ablation-results}
\end{figure*}

To visualize the outcomes of each module's ablation experiments, we exhibit its effect on the test set evaluation of DSIFN-CD, S2Looking and WHU-CD in Fig.~\ref{fig:ablation-results}. Although the prediction effect of each ablation experimental model for the bi-temporal images prediction in Fig.~\ref{fig:ablation-results} is mostly right. BD-MSA outperforms the other models in predicting the edges of the change region. The parts of the photography where BD-MSA outperforms the predictions of other models have been highlighted in yellow boxes.

In addition to ablation experiments on different modules, we also perform ablation studies on OFAM modules, specifically adding OFAM modules behind different stages in the backbone, as shown in Tables~\ref{tab:glam_dsifn}, \ref{tab:glam_s2looking} and~\ref{tab:glam_whu}. The results show that adding OFAM modules to all stages of the backbone has the greatest effect on the evaluation metrics, whereas OFAM-1 in the table is second-best in each evaluation index, which we hypothesize is due to the fact that the first stage of the backbone has the largest feature map, and the addition of OFAM modules can effectively aggregating information in the feature map, thus reducing the computational cost of the model.
\begin{table}[tb]
	\caption{The different stages in backbone are followed by the results of the OFAM ablation experiments on the DSIFN-CD test sets. We use different colors to indicate: \textcolor{red}{best}, \textcolor{blue}{second best}, and \textbf{third best}.}
	\label{tab:glam_dsifn}
	\centering
	\scalebox{0.8}{
	\begin{tabular}{c|cccc|cccc}
		\toprule
		\midrule
		Methods & stage1 & stage2 & stage3 & stage4 & F1 & Prec. & Rec. & IoU \\
		\midrule
		OFAM-1 & \checkmark &  &  &  & \textcolor{blue}{82.6} & \textcolor{blue}{87.68} & \textcolor{blue}{78.08} & \textcolor{blue}{70.36} \\
		OFAM-2 & \checkmark & \checkmark &  &  & \textbf{79.65} & \textbf{84.81} & \textbf{75.09} & \textbf{66.18} \\
		OFAM-3 & \checkmark & \checkmark & \checkmark &  & 76.32 & 83.73 & 70.11 & 61.71 \\
		OFAM-4 & \checkmark & \checkmark & \checkmark & \checkmark & \textcolor{red}{83.98} & \textcolor{red}{88.01} & \textcolor{red}{80.3} & \textcolor{red}{72.38} \\
		\midrule
		\bottomrule
	\end{tabular}
	}
\end{table}
\begin{table}[tb]
	\caption{The different stages in backbone are followed by the results of the OFAM ablation experiments on the S2Looking test sets. We use different colors to indicate: \textcolor{red}{best}, \textcolor{blue}{second best}, and \textbf{third best}.}
	\label{tab:glam_s2looking}
	\centering
	\scalebox{0.8}{
	\begin{tabular}{c|cccc|cccc}
		\toprule
		\midrule
		Methods & stage1 & stage2 & stage3 & stage4 & F1 & Prec. & Rec. & IoU \\
		\midrule
		OFAM-1 & \checkmark &  &  &  & \textcolor{blue}{63.66} & \textcolor{red}{74.29} & \textcolor{blue}{55.7} & \textcolor{blue}{46.7} \\
		OFAM-2 & \checkmark & \checkmark &  &  & 57.49 & 66.91 & 50.4 & 40.34 \\
		OFAM-3 & \checkmark & \checkmark & \checkmark &  & \textbf{60.01} & \textbf{67.59} & \textbf{53.96} & \textbf{42.86} \\
		OFAM-4 & \checkmark & \checkmark & \checkmark & \checkmark & \textcolor{red}{64.08} & \textcolor{blue}{70.44} & \textcolor{red}{58.77} & \textcolor{red}{47.14} \\
		\midrule
		\bottomrule
	\end{tabular}
	}
\end{table}
\begin{table}[tb]
	\caption{The different stages in backbone are followed by the results of the OFAM ablation experiments on the WHU-CD test sets. We use different colors to indicate: \textcolor{red}{best}, \textcolor{blue}{second best}, and \textbf{third best}.}
	\label{tab:glam_whu}
	\centering
	\scalebox{0.8}{
	\begin{tabular}{c|cccc|cccc}
		\toprule
		\midrule
		Methods & stage1 & stage2 & stage3 & stage4 & F1 & Prec. & Rec. & IoU \\
		\midrule
		OFAM-1 & \checkmark &  &  &  & \textcolor{blue}{92.6} & \textcolor{blue}{94.39} & \textcolor{blue}{90.88} & \textcolor{blue}{86.23} \\
		OFAM-2 & \checkmark & \checkmark &  &  & 89.8 & 90.94 & 88.69 & 81.48 \\
		OFAM-3 & \checkmark & \checkmark & \checkmark &  & \textbf{92.25} & \textcolor{red}{95.39} & \textbf{89.31} & \textbf{85.62} \\
		OFAM-4 & \checkmark & \checkmark & \checkmark & \checkmark & \textcolor{red}{93.41} & \textbf{94.3} & \textcolor{red}{92.53} & \textcolor{red}{87.63} \\
		\midrule
		\bottomrule
	\end{tabular}
	}
\end{table}

\begin{figure}
	\centering
	\includegraphics[width=0.9\linewidth]{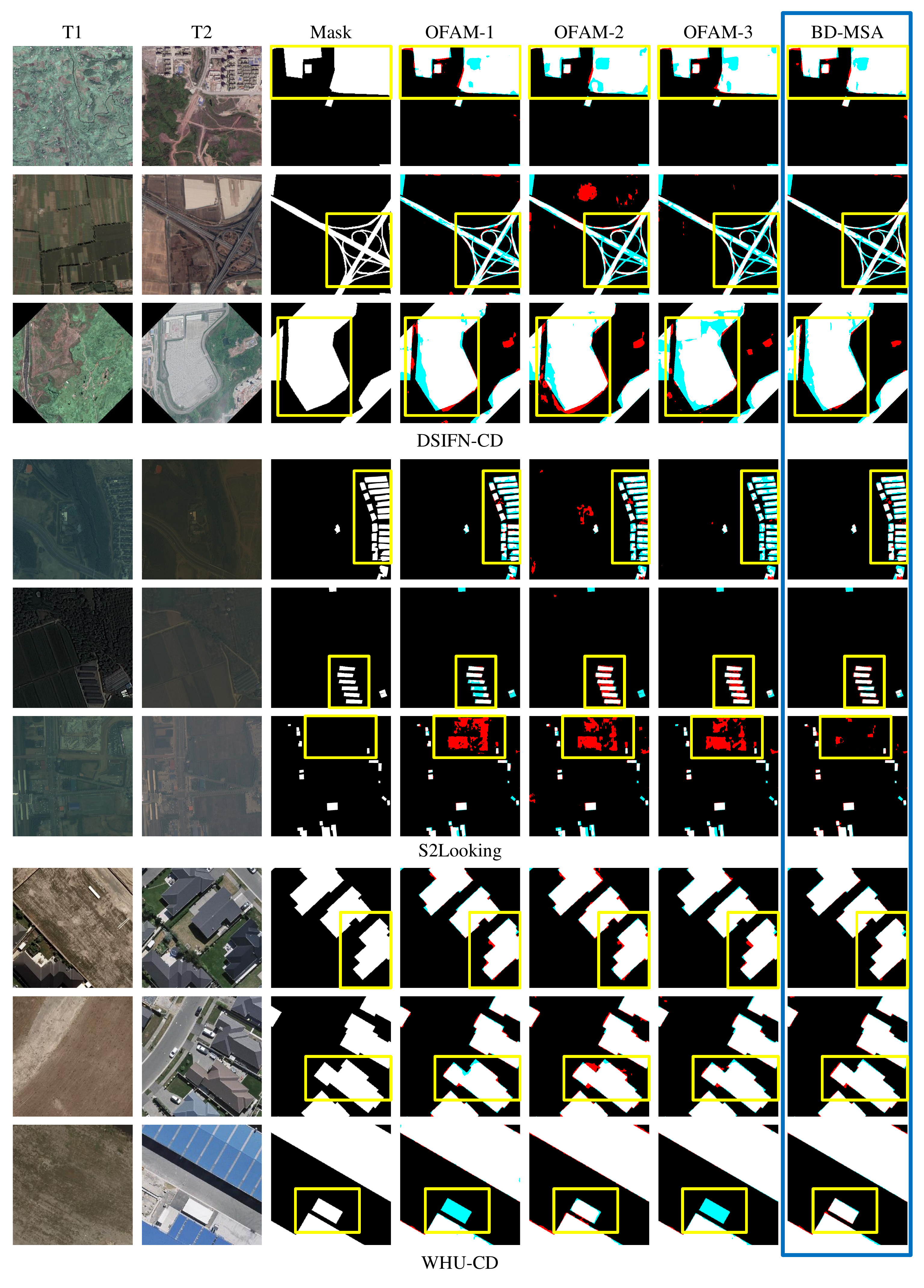}
	\caption{Visualization of the results of ablation experiments on DSIFN-CD, S2Looking and WHU-CD test sets for different stages followed by OFAM in backbone.}
	\label{fig:ablation-results-OFAM}
\end{figure}

Fig.~\ref{fig:ablation-results-OFAM} depicts the experimental outcomes of introducing OFAM behind various phases of the backbone. In general, each model achieves better prediction results, but BD-MSA outperforms the other models in the subtle aspects shown in the figure with yellow boxes, such as edge detection, which is more accurate and can separate buildings with tight layouts very well.

\subsection{Feature Map Visualization}
\begin{figure}
	\centering
	\includegraphics[width=0.8\linewidth]{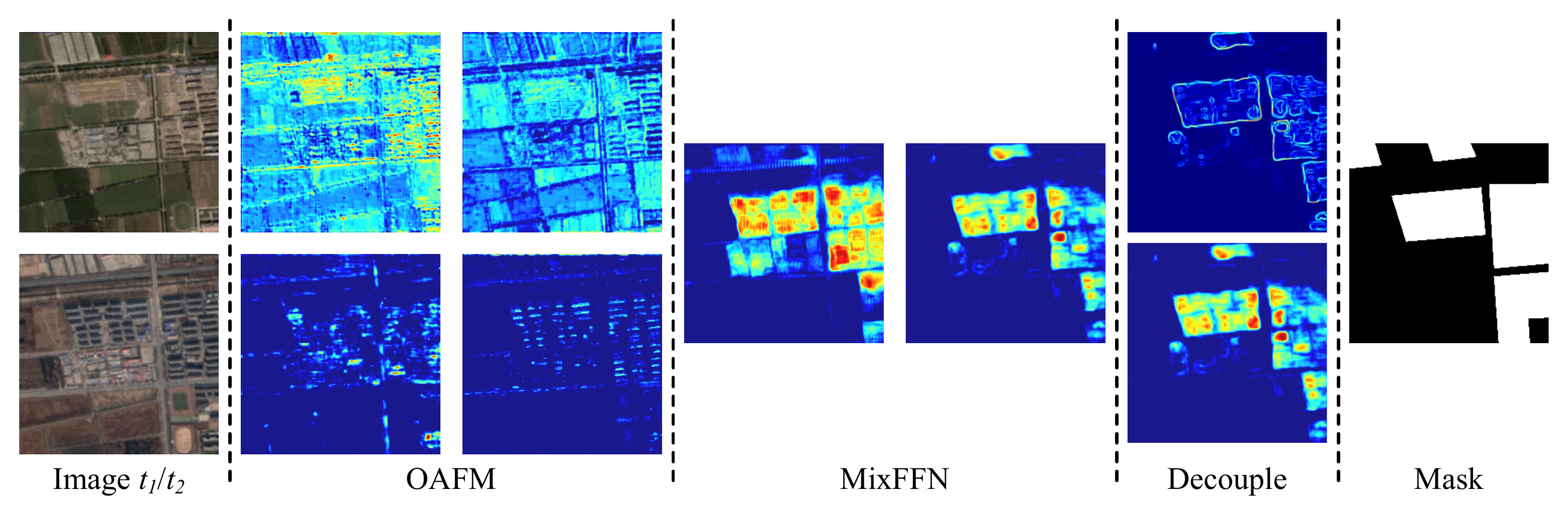}
	\caption{Visualization of heat maps generated by some modules.}
	\label{fig:heat_map}
\end{figure}

To investigate whether the modules in this paper's model are able to aggregate semantic information in the prediction process for bi-temporal images, we used Grad-CAM~\cite{selvaraju2017grad} to view some of the feature layers in BD-MSA, and the results are shown in Fig.~\ref{fig:heat_map}.

From left to right, the figure is divided into five sections: the original bi-temporal images, feature maps before and after OFAM for stage 1, feature maps before and after MixFFN, boundary and body feature maps generated by Decouple Module, and change labels.

The figure clearly shows that the OFAM Module can transfer the weight in the feature map from the unimportant road part to the more important building part; MixFFN can focus the features on the changing region while reducing the weight of the non-changing region; and Decouple Module can effectively decouple the feature map and extract the edge features.

\section{Discussion}
\label{sec:discussion}
In this section, we discuss the following three issues: the effect of different datasets on the experimental results, how different hyper-parameters affect the model performance and how semi-supervised learning methods affect the overall performance.
\subsection{Effect of training set on experimental results}
An essential factor influencing the experimental outcomes during model training is the quality of the dataset. When we conducted the experiments, we discovered that the final test accuracy varied significantly between datasets. For instance, the IoU on the test set in WHU-CD reached 87.63\%, while the IoU on the test set in S2Looking was only 47.14\%. Based on our conjectures, we determined that one of the causes of this phenomenon was the datasets' excessively variable proportion of positive and negative samples. To address this, we counted the number of pixels in each dataset as well as the percentage of positive samples overall, as indicated in Table~\ref{tab:positive ratio}.

The findings demonstrate that while the proportion of positive sample pixels in S2Looking is very low at 1.27\% of total pixels, it is also low in WHU-CD, at 4.26\%, significantly lower than in DSIFN-CD, which has a proportion of 35.03\%. Through an examination of the images in the dataset, we discovered that while the percentage of positive samples in WHU-CD is significantly lower than in DSIFN-CD, the reason for this phenomenon is that the majority of WHU-CD's areas remain unchanged, and in the images that have changed, the altered areas are all buildings, all of which have very regular shapes. The models obtain a reasonably decent level of accuracy in this dataset since the lighting, contrast, and shooting angle are all extremely perfect.
\begin{table}[htbp]
	\caption{The ratio of positive and negative pixel samples in different datasets}
	\label{tab:positive ratio}
	\centering
	\begin{tabular}{c|ccc}
		\toprule
		\midrule
		Datasets & positive sample & negative sample & ratio \\
		\midrule
		DSIFN-CD & 361819124 & 671028236 & 35.03\% \\
		S2Looking & 66552990 & 5176327010 & 1.27\% \\
		WHU-CD & 21442501 & 481873976 & 4.26\% \\
		\midrule
		\bottomrule
	\end{tabular}
\end{table}

\subsection{Effect of different Hyperparameters on model performance}
In every experiment in this work, we use AdamW as the optimizer and BCE Loss as the loss function. Since we are using the Open-CD development kit, we utilize PolyLR, the default learning rate strategy, for learning rates. We test the model on all three of the publicly accessible datasets for which the aforementioned hyperparameters are modified in order to investigate the effects of various hyperparameters on the model's performance.

The various methods for setting hyperparameters are as follows:
\begin{itemize}
	\item[(1)] $\lambda_{1}$: BCE Loss + AdamW + PolyLR.
	\item[(2)] $\lambda_{2}$: Dice Loss + AdamW + PolyLR.
	\item[(3)] $\lambda_{3}$: BCE Loss + SGD + PolyLR.
	\item[(4)] $\lambda_{4}$: BCE Loss + AdamW + StepLR.
\end{itemize}

\begin{figure}
	\centering
	\includegraphics[width=1\linewidth]{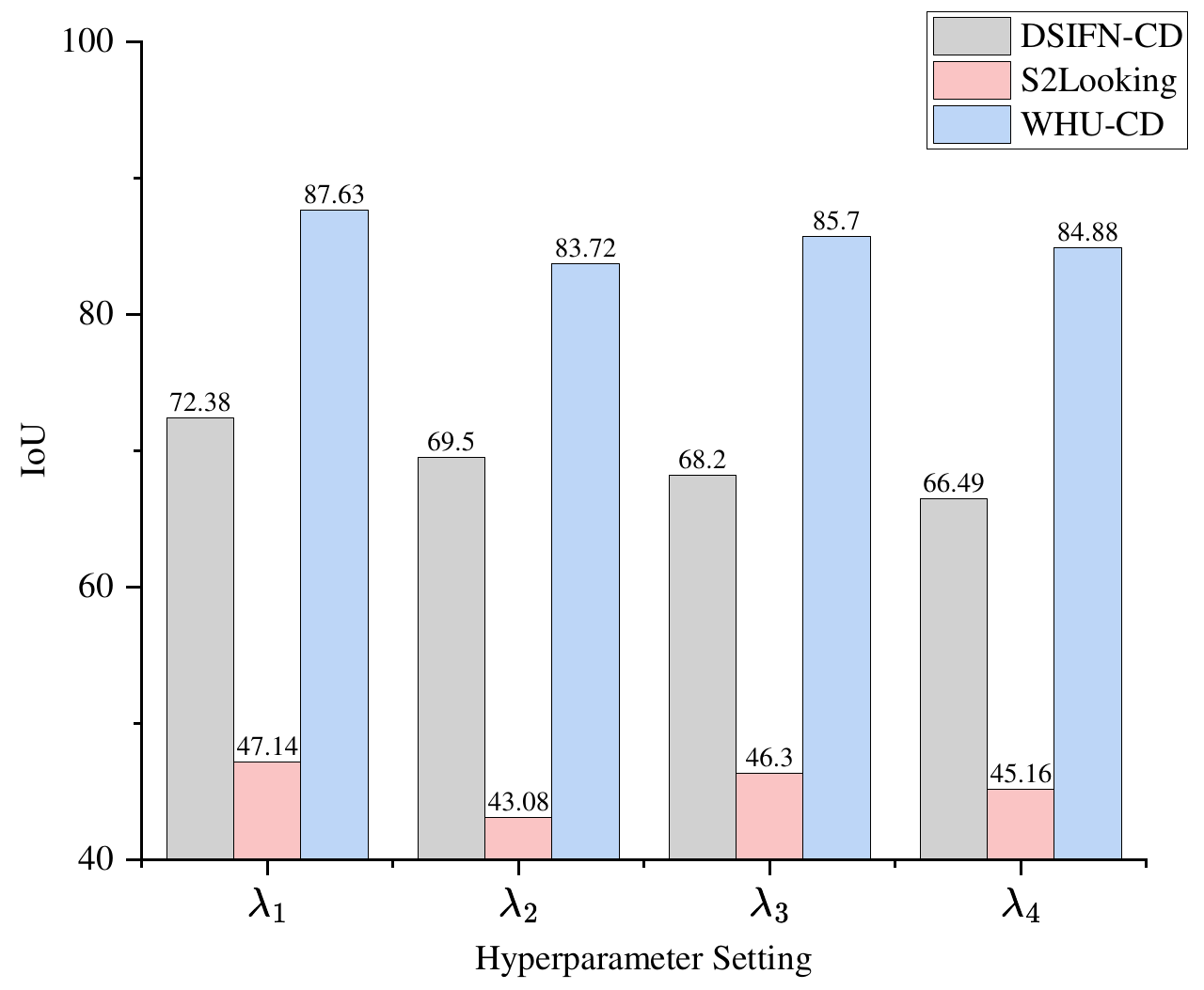}
	\caption{Experimental results for different hyperparameter settings on different datasets.}
	\label{fig:hyperparameter}
\end{figure}

Fig.~\ref{fig:hyperparameter} displays the outcomes of the experiment. The IoU of evaluation metrics on each dataset varies somewhat depending on the hyperparameter settings used. By comparing the data in the image, it can be seen that the hyperparameter setting of $\lambda_{1}$ (BCE Loss + AdamW + PolyLR) yields the greatest IoU across all datasets.

\subsection{Impact of Semi-supervised Learning on Model Performance}
This study proposes the BD-MSA model, which is mainly based on supervised learning and necessitates a huge quantity of labeled data. It remains challenging to gather a significant number of high-quality bi-temporal remote sensing images of the same region in the real world, despite the fact that the experimental setting used in this study can accommodate a sizable number of datasets for training.

To address the aforementioned problems, this paper simulates the semi-supervised learning approach and investigates how it influences the model's overall performance by randomly sampling the training sets from each public dataset in proportions of 5\%, 10\%, 20\%, and 40\%, respectively. The training sets are then configured with the same hyper-parameter settings. Table~\ref{tab:semi supervised} presents the experimental outcomes.

It makes logical sense that when the sample ratio rises, as demonstrated by the experimental findings, the model's various assessment metrics across datasets increase. Remarkably, BD-MSA attains relatively good assessment metrics on S2Looking and WHU-CD upon reaching a sample ratio of 40\%.

Since this paper's methodology is fundamentally a supervised learning strategy, change detection in semi-supervised learning will necessarily be less accurate. However, utilizing less than half of the training set data volume, a relatively good accuracy was obtained, a phenomena that captures our interest and can be focused on semi-supervised learning in future work.
\begin{table*}[htbp]
	\caption{Experimental results of semi-supervised learning of BD-MSA with different labeled Ratio on different datasets, We use different colors to indicate: \textcolor{red}{best}, \textcolor{blue}{second best}, and \textbf{third best}.}
	\label{tab:semi supervised}
	\centering
	\begin{tabular}{c|cccc|cccc|cccc}
		\toprule
		\midrule
		\multirow{2}{*}{Labeled Ratio} & \multicolumn{4}{c|}{DSIFN-CD} & \multicolumn{4}{c|}{S2Looking} & \multicolumn{4}{c}{WHU-CD} \\
		& F1 & Prec. & Rec. & IoU & F1 & Prec. & Rec. & IoU & F1 & Prec. & Rec. & IoU\\
		\midrule
		5\% & 50.33 & 68.03 & 39.94 & 33.63 & 55.07 & 58.22 & 52.25 & 38.0 & 74.74 & \textbf{83.3} & 67.78 & 59.67 \\
		10\% & 55.86 & 69.11 & 46.87 & 38.75 & 56.31 & 61.37 & 52.02 & 39.19 & 80.11 & 82.44 & 77.91 & 66.82 \\
		20\% & \textbf{57.95} & \textcolor{blue}{75.75} & \textbf{46.93} & \textbf{40.8} & \textbf{58.73} & \textbf{64.6} & \textbf{53.84} & \textbf{41.58} & \textbf{83.94} & 79.22 & \textbf{89.27} & \textbf{72.33} \\
		40\% & \textcolor{blue}{69.59} & \textbf{69.38} & \textcolor{blue}{69.8} & \textcolor{blue}{53.36} & \textcolor{blue}{60.73} & \textcolor{blue}{66.25} & \textcolor{blue}{56.06} & \textcolor{blue}{43.61} & \textcolor{blue}{87.03} & \textcolor{blue}{84.6} & \textcolor{blue}{89.6} & \textcolor{blue}{77.04} \\
		100\% & \textcolor{red}{83.98} & \textcolor{red}{88.01} & \textcolor{red}{80.3} & \textcolor{red}{72.38} & \textcolor{red}{64.08} & \textcolor{red}{70.44} & \textcolor{red}{58.77} & \textcolor{red}{47.14} & \textcolor{red}{93.41} & \textcolor{red}{94.3} & \textcolor{red}{92.53} & \textcolor{red}{87.63} \\
		\midrule
		\bottomrule
	\end{tabular}
\end{table*}

\section{Conclusion}
\label{sec:conclusion}
In this study, we suggested a novel approach for RSCD called BD-MSA. In the training and prediction phase, the approach can combine global and local information in both channel and spatial dimensions, as well as decouple the main body of the change region and the edges of the feature maps. The experimental results suggest that the technique in this research outperforms previous models on the public datasets DSIFN-CD, S2Looking and WHU-CD in terms of SOTA performance. We further demonstrate, through a series of ablation experiments, that all modules in this study are superior to the baseline.

We will continue to investigate the following aspects in the future: 1) The method in this paper has only been validated on three public datasets, DSIFN-CD, S2Looking and WHU-CD, and it will be validated on more public datasets in the future; 2) The method in this paper is essentially a supervised learning method, and we hope to explore unsupervised learning methods for application to tasks such as remote sensing image change detection and more domain migration in future work.

\section*{Acknowledgment}
This research was supported by National Natural Science Foundations of China (No. 42261078), the Jiangxi Provincial Key R\&D Program (Grant number20223BBE51030) and the Science and Technology Research Project of Jiangxi Bureau of Geology(Grant number 2022JXDZKJKY08) and the Open Research Fund of Key Laboratory of Mine Environmental Monitoring and Improving around Poyang Lake of Ministry of Natural Resources(MEMI-2021-2022-31) and the Graduate Innovative Special Fund Projects of Jiangxi Province(YC2023-S556).

\bibliographystyle{IEEEtran}
\bibliography{Inference}

\vfill
	
\end{document}